\documentclass[letterpaper,journal]{IEEEtran}
\usepackage{amssymb,amsfonts}
\usepackage{graphicx}
\usepackage{textcomp}
\usepackage{amsmath}
\usepackage{bm}
\usepackage{makecell}
\usepackage{booktabs}
\usepackage{multirow}
\usepackage{tabularx}
\usepackage{ragged2e}
\usepackage{placeins}
\usepackage{xcolor}
\usepackage{cite}
\usepackage{url}
\def\BibTeX{{\rm B\kern-.05em{\sc i\kern-.025em b}\kern-.08em
    T\kern-.1667em\lower.7ex\hbox{E}\kern-.125emX}}
\usepackage{array}

\begin{document}
\title{Agentic Semantic Sensing for Resource-Adaptive AI-RAN}

\author{Zhongqin Wang, \IEEEmembership{Member, IEEE},
		Xiaoqi Zhang, \IEEEmembership{Student Member, IEEE},
		Nan Yang, \IEEEmembership{Member, IEEE}, \\
		Kai Wu, \IEEEmembership{Member, IEEE},
		J. Andrew Zhang, \IEEEmembership{Senior Member, IEEE},
		Y. Jay Guo, \IEEEmembership{Life Fellow, IEEE}

\IEEEcompsocitemizethanks{
\IEEEcompsocthanksitem Zhongqin Wang, Xiaoqi Zhang, Kai Wu, J. Andrew Zhang (corresponding author), and Y. Jay Guo are with the Global Big Data Technologies Centre, University of Technology Sydney, Sydney, NSW 2007, Australia. E-mail: \{zhongqin.wang, kai.wu, andrew.zhang, jay.guo\}@uts.edu.au; xiaoqi.zhang@student.uts.edu.au.
\IEEEcompsocthanksitem Nan Yang is with the School of Science, Engineering and Digital Technologies, University of Southern Queensland, Toowoomba, QLD 4350, Australia, and the School of Electrical and Computer Engineering, The University of Sydney, Sydney, NSW 2006, Australia. E-mail: nan.yang@unisq.edu.au.
}
}
\maketitle

\begin{abstract}
Semantic sensing (SemS) acquires task-relevant information rather than reconstructing complete physical information. Existing SemS formulations typically operate open loop: sensing configurations and observation schedules are fixed before inference and cannot respond to evolving task-level evidence. We propose Agentic SemS, a closed-loop framework for AI-enabled radio access networks (AI-RANs) that controls sensing within a communication-feasible profile set. A profile-conditioned causal Transformer updates the semantic belief from streaming observations, while key--value caching enables efficient state updates across profile changes without repeatedly processing the complete history. A semantic utility network estimates the task-level benefit of acquiring the next observation block under each feasible profile after accounting for sensing cost. The resulting continuation utilities jointly support next-profile selection and semantic early exit, adapting sensing configuration and duration to evolving evidence. The expected semantic gain is further related to conditional mutual information, providing a value-of-information interpretation of continued online sensing. Experiments on Widar3.0 with six emulated sensing profiles show that, in comparison with full-sequence High, the resource-efficient Agentic setting reduces normalized cumulative sensing cost by 25.33\% while achieving 85.79\% Macro-F1. At the same utility checkpoint, semantic early exit provides a further 12.35\% cost reduction over adaptive sensing without early exit, with a 0.97-percentage-point Macro-F1 decrease.
\end{abstract}


\section{Introduction}
\IEEEPARstart{N}{ext-generation} wireless networks are expected to provide connectivity, perceive their environment, and adapt operation to evolving task requirements. Integrating artificial intelligence (AI) into the radio access network (RAN), known as AI-RAN~\cite{kundu2026airan,polese2026beyond}, enables learning-driven sensing, inference, and resource control, while data-driven architectures support adaptive resource management and service provisioning~\cite{fang2019datadriven}. Emerging agentic AI closes the perception--decision--action loop by updating task-relevant knowledge from observations, evaluating feasible actions, and using selected actions to shape subsequent environmental interactions. Integrated sensing and communication (ISAC) extends this capability by allowing sensing and communication to share spectrum, signals, energy, and hardware resources~\cite{9296833,zhang2021overview,wu2022joint,liu2022integrated,wu2025ISAC}. Within such systems, information requirements for reliable inference can vary with task progress, environmental dynamics, and wireless conditions. This gives rise to two coupled resource-control decisions: how the sensing configuration should be adapted when additional information is required, and whether further sensing should be performed at all. These decisions determine the resource cost of continued acquisition and the duration over which sensing resources are consumed.

Conventional wireless sensing includes physical-parameter estimation and learning-based task inference. Classical signal-processing approaches estimate physical parameters such as propagation delays, Doppler shifts, angles of arrival (AoAs), and path gains~\cite{wu2024bistatic,li2024otfs}. Learning-based methods map received signal strength indicator (RSSI)~\cite{wang2026rethinking}, channel state information (CSI)~\cite{wang2026multipoint}, and Doppler features~\cite{wang2025siso} directly to task labels or task-specific estimates. Recent work has explored technology-agnostic recognition across heterogeneous radio-frequency (RF) modalities~\cite{yang2023tarf} and resource-adaptive inference under varying channel and computation budgets~\cite{waqas2025snr}. These methods can achieve strong inference, but their sensing configurations and observation schedules are typically fixed before inference. Such open-loop operation cannot respond to changes in the informativeness of newly acquired observations, task uncertainty, environmental dynamics, or wireless conditions. Consequently, a fixed sensing configuration may expend excessive resources when lower-cost observations are already sufficient, while a fixed observation horizon may continue acquisition after reliable inference has become possible or terminate before enough task-relevant evidence has been collected. Addressing this mismatch requires sensing decisions to be guided by task-relevant information rather than a predetermined strategy.

Semantic sensing (SemS)~\cite{zhang2026semantic} provides such a task-oriented perspective by focusing sensing on information relevant to downstream inference rather than complete physical-information reconstruction. This direction is closely related to semantic communication, which moves beyond exact bit reconstruction and emphasizes the delivery of task-relevant meaning~\cite{lu2024semantics,sagduyu2024joint}. Token-based semantic communication further exploits cross-modal context to improve communication efficiency~\cite{qiao2025token}, while related task-oriented work in integrated sensing, communication, and computation allocates radio and computing resources according to edge-learning or inference performance rather than reconstruction fidelity~\cite{xing2023taskoriented}. Existing SemS formulations establish the principle of task-relevant information acquisition, but sensing configurations and observation procedures are generally determined before inference. As a result, the task-level information revealed during sensing does not directly influence how subsequent observations are acquired. Closing this feedback loop would allow the evolving inference state to adapt the sensing profile and observation duration according to the information required for reliable inference, thereby reducing unnecessary sensing-side radio and processing resource expenditure.

Recent studies on feedback-driven ISAC and agentic AI examine closed-loop control of predictive beamforming, target-aware transmission, spectrum allocation, power control, and communication--sensing resource management~\cite{du2023dynamic,zhang2024predictive,xiao2025toward,zhang2026towardagentic,jiang2026large,xie2025agenticisac}. Adaptive-interval CSI estimation reduces measurement overhead by deciding whether communication and radar CSI should be re-estimated or predicted from temporal correlation~\cite{chen2024intermittent}, while feedback-aware cognitive-radio designs jointly control spectrum sensing and access under resource constraints~\cite{elshafie2012optimal}. These approaches use channel states, target dynamics, traffic conditions, or network-level objectives to drive adaptation. In contrast, they do not use the evolving task-level inference state as feedback for subsequent sensing. This leaves open the question of how semantic inference itself can guide which sensing profile should be used next and, when appropriate, whether further sensing is necessary.

In this work, we propose \emph{Agentic SemS}, a closed-loop framework that uses evolving semantic information to control subsequent sensing in AI-RANs. A sensing profile specifies the observation quality, signal representation, temporal configuration, and sensing-resource usage of an observation block, while the RAN defines the set of profiles that satisfy the underlying communication requirements. A profile-conditioned causal Transformer~\cite{vaswani2017attention} processes streaming observations and incrementally updates the semantic belief, which represents the current task-level inference state. Key--value (KV) caching preserves the accumulated observation context across profile transitions, enabling efficient incremental inference without repeatedly processing the complete observation history~\cite{pope2023efficiently}. At each decision epoch, a semantic utility network evaluates the feasible sensing profiles according to the current semantic state and their predicted contribution to subsequent inference. The agent then selects the profile for the next observation block, allowing the sensing configuration and associated resource usage to evolve with the task-level evidence. Semantic early exit provides an additional duration-control mechanism, allowing the agent to terminate sensing when further acquisition is predicted to provide limited benefit. Repeating this belief--utility--action cycle establishes a closed feedback loop between semantic inference and sensing-resource adaptation.

Our main contributions are summarized as follows.
\begin{itemize}

\item We propose Agentic SemS, a closed-loop semantic sensing framework that uses the evolving semantic belief as feedback for subsequent acquisition. Within a communication-feasible profile set, a continuation-utility formulation evaluates the predicted task-level benefit and sensing-resource cost of further observation and supports both next-profile selection and semantic early exit. The expected semantic gain is linked to conditional mutual information, providing a value-of-information (VoI) interpretation of continued sensing.

\item We provide a reference implementation of Agentic SemS that realizes the framework as a streaming sensing system. It covers profile-dependent sensing front ends, the architectures and training procedures of the semantic belief and utility networks, and KV-cached online operation for profile adaptation and semantic early exit, while allowing its front ends and profile context to adapt across sensing modalities and configurations in diverse deployments.

\item We evaluate Agentic SemS on Widar3.0~\cite{zhang2021widar3} with six emulated sensing profiles and controlled comparisons of profile adaptation and semantic early exit.\footnote{The source code will be made publicly available upon acceptance.} In comparison with full-sequence High, the resource-efficient Agentic setting reduces normalized cumulative sensing cost by 25.33\% while achieving 85.79\% Macro-F1, demonstrating that the two control mechanisms can reduce sensing-resource usage while maintaining task performance.
\end{itemize}

The paper is organized as follows. Section~II reviews SemS and motivates its closed-loop extension. Section~III presents the Agentic SemS framework. Section~IV describes the system design and implementation, while Section~V presents the experimental setup and results. Section~VI concludes the paper.

\begin{figure*}[t]
\centering
\includegraphics[width=0.95\textwidth]{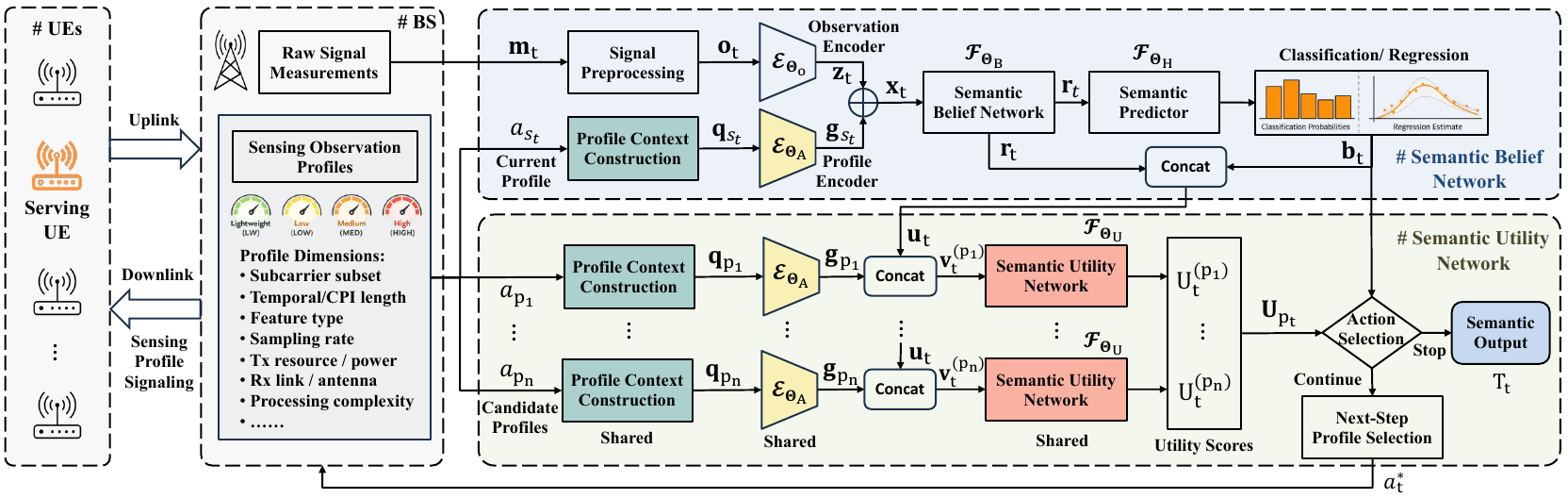}
\caption{Closed-loop architecture of Agentic SemS. The belief network updates the semantic state from streaming observations, while the utility network scores feasible profiles. The predicted utilities jointly support semantic early exit and, when sensing continues, selection of the next feasible profile.}
\label{fig:agentic_framework}
\vspace{-1.5em}
\end{figure*}

\section{Preliminaries and Motivation}
\label{sec:preliminaries}
This section briefly reviews the open-loop SemS formulation and motivates its extension to closed-loop Agentic SemS.

\subsection{Semantic Sensing Model}
\label{subsec:sems_model}
Semantic sensing (SemS)~\cite{zhang2026semantic} provides a task-oriented sensing paradigm that focuses information acquisition on downstream inference rather than complete environmental reconstruction. Let $\mathcal{A}=\{a_1,\ldots,a_{N_{\rm p}}\}$ denote the sensing-profile space, where each profile specifies a sensing configuration and its associated resource usage. In an AI-RAN system, communication-service requirements constrain the sensing configurations that can be used. We represent the resulting feasible profile set as
\begin{equation}
\mathcal{A}^{\rm com}
=
\left\{
a_p\in\mathcal{A}:
Q_{\rm com}(a_p)
\geq
Q_{\rm com}^{\min}
\right\},
\label{eq:communication_feasible_resources}
\end{equation}
where $Q_{\rm com}(a_p)$ is the communication quality of profile $a_p$, and $Q_{\rm com}^{\min}$ is the minimum communication-quality requirement. Accordingly, $\mathcal{A}^{\rm com}$ is the communication-feasible profile set considered throughout the framework and is supplied by the underlying RAN to the sensing controller.

A generic profile-dependent SemS acquisition model is
\begin{equation}
\mathbf{x}_{s}^{(p)}
=
\mathcal{F}_{\Theta_{\rm Tx}}
\left(
\mathbf{x}_{\rm ref};a_p
\right),
\qquad
a_p\in\mathcal{A}^{\rm com},
\label{eq:semantic_encoder}
\end{equation}
where $\mathbf{x}_{\rm ref}$ is the reference signal, $\Theta_{\rm Tx}$ contains encoder parameters, and $a_p$ identifies the active profile. A profile may specify subcarrier usage, sampling rate, signal quality, per-block temporal configuration, and signal representation.

The sensing signal interacts with the physical environment and propagation channel to produce
\begin{equation}
\mathbf{y}_{s}^{(p)}
=
\Psi_{\rm ch}
\left(
S,\mathbf{x}_{s}^{(p)}
\right)
+
\mathbf{n}_{s},
\label{eq:semantic_observation}
\end{equation}
where $S$ denotes the latent physical state, $\Psi_{\rm ch}(\cdot)$ represents the sensing-channel mapping, and $\mathbf{n}_{s}$ denotes aggregate noise. The semantic decoder directly estimates the task output as
\begin{equation}
\hat{T}
=
\mathcal{F}_{\Theta_{\rm Rx}}
\left(
\mathbf{y}_{s}^{(p)}
\right),
\label{eq:semantic_decoder}
\end{equation}
without reconstructing task-irrelevant physical information.

\subsection{From Open-Loop SemS to Closed-Loop Agentic SemS}
\label{subsec:from_sems_to_agentic}
The model in~\eqref{eq:semantic_encoder}--\eqref{eq:semantic_decoder} captures profile-dependent task-oriented sensing but fixes the sensing configuration and observation procedure, preventing feedback from the evolving task-level inference state. Consequently, information revealed during sensing does not directly control subsequent acquisition or whether sensing should continue.
As sensing proceeds, the value of observations changes with accumulated evidence, environmental dynamics, and wireless conditions. A predetermined profile may waste resources when lower-cost observations suffice, while a fixed schedule cannot adapt duration to evolving inference needs. Agentic SemS closes this loop by using the semantic belief to guide sensing. At each epoch, the agent selects a feasible profile as evidence accumulates. Semantic early exit terminates sensing when further acquisition has low predicted value, jointly adapting sensing configuration and duration to task requirements.

\section{Agentic Semantic Sensing Framework}
\label{sec:agentic_architecture}
This section presents how Agentic SemS represents the evolving task-level information, evaluates feasible sensing profiles, and controls subsequent sensing.

\subsection{System Overview}
\label{subsec:system_overview}

\subsubsection{Closed-Loop Semantic Sensing Task}
\label{subsubsec:semantic_task_definition}
We consider a semantic sensing task with target $T$. For classification, $T=y\in\{1,\ldots,C\}$, whereas for regression, $T=\mathbf{y}\in\mathbb{R}^{D_{\rm y}}$, where $C$ and $D_{\rm y}$ denote the number of classes and the output dimension, respectively. As shown in Fig.~\ref{fig:agentic_framework}, Agentic SemS processes observations under the active sensing profile and updates the task-level inference state. At each decision epoch, the agent selects a feasible sensing profile for the next observation block. Semantic early exit adapts sensing duration according to the predicted value of further acquisition. The resulting belief--utility--action sequence closes the loop, so each sensing decision affects subsequent observations.

\subsubsection{Sensing Profiles and Feasible Actions}
\label{subsubsec:feasible_sensing_profiles}

Using the sensing-profile space $\mathcal{A}$ and communication-feasible subset $\mathcal{A}^{\rm com}$ defined in Section~\ref{sec:preliminaries}, each profile specifies an operating point through its observation, configuration, and resource attributes. Let $s_t$ denote the profile active at observation step $t$, and let $\chi_t(a_p,a_{s_t})\in\{0,1\}$ indicate whether profile $a_p$ is available from the current operating state. The feasible continuation profiles at decision epoch $t$ are
\begin{equation}
\mathcal{A}_t^{\rm feas}
=
\left\{
a_p\in\mathcal{A}^{\rm com}
:
\chi_t(a_p,a_{s_t})=1
\right\},
\label{eq:feasible_profile_set}
\end{equation}
where $\chi_t(\cdot)$ captures sensing-hardware availability and reconfiguration constraints at decision epoch $t$.

\subsubsection{Semantic Belief and State}
\label{subsubsec:semantic_belief_state}
At observation step $t$, active profile $a_{s_t}$ is described by context $\mathbf{q}_{s_t}$ and used to acquire observation $\mathbf{o}_t$. The belief network processes accumulated observations with corresponding profile contexts and produces a latent representation $\mathbf{r}_t$ and a task-dependent semantic belief $\mathbf{b}_t$. For classification, $\mathbf{b}_t$ represents the posterior distribution over the semantic classes; for regression, it may represent the predictive estimate together with its uncertainty. Together, $\mathbf{r}_t$ and $\mathbf{b}_t$ constitute the semantic state $\mathbf{u}_t$, which combines accumulated task-relevant information with the current inference to guide subsequent profile selection and stopping.

\subsubsection{Joint Profile Selection and Semantic Early Exit}
\label{subsubsec:continuation_utility_selection}
For each $a_p\in\mathcal{A}_t^{\rm feas}$, the encoder maps its context $\mathbf{q}_p$ to an embedding $\mathbf{g}_p$. The semantic utility network combines $\mathbf{g}_p$ with the semantic state $\mathbf{u}_t$ to estimate the utility of acquiring the next observation block under profile $a_p$. The predicted utilities support two coupled decisions: selecting a sensing profile for the next observation block and determining whether further sensing is worthwhile. If sensing continues, the selected profile is applied to the next observation block; otherwise, sensing terminates and the current task prediction is returned. Each new block updates the semantic state before subsequent profile-selection and stopping decisions during online operation.

\subsection{Semantic Belief Network}
\label{subsec:semantic_belief_network}
The belief network is the perception and state-update module of Agentic SemS and uses each coherent processing interval (CPI) as an observation unit. A CPI aggregates a short sequence of consecutive measurements, providing sufficient temporal support for extracting motion-sensitive spectral or statistical features, such as Doppler features, while preserving the temporal resolution required for online inference. The resulting CPI-level observation provides a common sequential input unit across sensing profiles. At each CPI, the belief network processes the observation together with its active-profile context and incrementally updates the latent representation $\mathbf{r}_t$ and semantic belief $\mathbf{b}_t$. The resulting state supplies task-level information for subsequent sensing decisions.

\subsubsection{CPI-Level Input Representation}
\label{subsubsec:cpi_input_representation}
At CPI $t$, the profile-dependent preprocessing module receives measurements $\mathbf{m}_t$ acquired under sensing profile $a_{s_t}$. Depending on the sensing interface and active profile, $\mathbf{m}_t$ may comprise raw in-phase/quadrature (I/Q) samples, channel-response measurements such as CSI, or compact radio indicators such as RSSI, reference signal received power (RSRP), reference signal received quality (RSRQ), and signal-to-interference-plus-noise ratio (SINR). Their modality, structure, and dimension may vary across profiles according to the sensing hardware and the information exposed by the active configuration. The sensing profile therefore determines which information is available and the resources required to acquire it.

\subsubsection{Signal Preprocessing and Observation Encoding}
\label{subsubsec:profile_dependent_signal_preprocessing}
Profile-dependent preprocessing of $\mathbf{m}_t$ under $a_{s_t}$ suppresses measurement artifacts and produces the CPI observation
\begin{equation}
\mathbf{o}_t
=
\mathcal{G}_{\rm o}
\left(
\mathbf{m}_t;
a_{s_t}
\right)
\in
\mathbb{R}^{N_{\rm obs}},
\label{eq:cpi_observation_preprocessing}
\end{equation}
where $\mathcal{G}_{\rm o}(\cdot)$ denotes the profile-dependent preprocessing and feature-extraction function. For example, CSI measurements may be affected by timing offsets, residual carrier-frequency offsets, receiver-chain phase distortions, automatic-gain-control variations, static multipath components, and interference. Although different profiles may provide different measurements and acquisition settings, their outputs are mapped to the common dimension $N_{\rm obs}$.

The resulting observation is mapped to a common embedding through a learnable observation encoder:
\begin{equation}
\mathbf{z}_t
=
\mathcal{E}_{\theta_{\rm o}}(\mathbf{o}_t)
\in
\mathbb{R}^{N_{\rm B}},
\label{eq:observation_encoding}
\end{equation}
where $\mathcal{E}_{\theta_{\rm o}}(\cdot)$ is an $N_{\rm B}$-dimensional observation encoder.

\subsubsection{Profile-Context Encoding}
\label{subsubsec:profile_context_construction}

Each sensing profile $a_p$ is associated with a context vector $\mathbf{q}_p$ that provides a compact numerical description of its sensing configuration and resource requirements, rather than the sensed data itself. Depending on the system, this context may include the sensing modality, sampling rate, observation quality, and temporal configuration; the sensing-side use of subcarriers, bandwidth, antennas, or beams; and the associated normalized sensing-resource cost. This context enables the belief network to interpret observations according to how they were acquired and allows the utility network to evaluate candidate profiles under their respective sensing conditions and resource requirements. The profile context is encoded as
\begin{equation}
\mathbf{g}_p
=
\mathcal{E}_{\theta_{\rm A}}
\left(
\mathbf{q}_p
\right)
\in
\mathbb{R}^{N_{\rm B}},
\label{eq:active_profile_context_embedding}
\end{equation}
where $\mathcal{E}_{\theta_{\rm A}}(\cdot)$ denotes the profile encoder. The same encoder represents the active profile for semantic belief updates and the candidate profiles for subsequent utility evaluation.

\subsubsection{Semantic Belief Update and Prediction}
\label{subsubsec:semantic_belief_update}
At CPI $t$, the observation embedding, active-profile embedding, and positional embedding are fused through element-wise addition:
\begin{equation}
\mathbf{x}_t
=
\mathbf{z}_t
+
\mathbf{g}_{s_t}
+
\mathbf{p}_t
\in
\mathbb{R}^{N_{\rm B}},
\label{eq:profile_conditioned_cpi_input}
\end{equation}
where $\mathbf{p}_t$ denotes the positional embedding. This construction conditions each observation on its acquisition profile.

For online sensing decisions, the belief model uses causal attention, such that the representation at CPI $t$ is computed only from the current and preceding profile-conditioned inputs, without access to future observations:
\begin{equation}
\mathbf{r}_t
=
\mathcal{F}_{\theta_{\rm B}}
\left(
\mathbf{x}_{1:t}
\right),
\label{eq:causal_semantic_belief_update}
\end{equation}
where $\mathbf{r}_t$ denotes the latent semantic representation at CPI $t$. Because each input $\mathbf{x}_t$ includes the corresponding active-profile context, $\mathbf{r}_t$ encodes both the accumulated observations and the sensing conditions under which they were acquired.

For classification, the semantic belief is obtained as
\begin{equation}
\mathbf{b}_t
=
\operatorname{softmax}
\left(
\mathbf{W}_{\rm cls}\mathbf{r}_t
+
\boldsymbol{\beta}_{\rm cls}
\right)
\in
\mathbb{R}^{C},
\label{eq:classification_semantic_belief}
\end{equation}
where $\mathbf{W}_{\rm cls}$ and $\boldsymbol{\beta}_{\rm cls}$ are learnable parameters, and $\mathbf{b}_t=[b_{t,1},\ldots,b_{t,C}]^{\mathsf T}$ contains the class-posterior probabilities. For regression, the same latent representation can be coupled with a task-specific prediction head that outputs the estimate and associated uncertainty. \textit{For clarity, the following formulation uses classification to illustrate the framework, although Agentic SemS is not limited to classification tasks.} The resulting $\mathbf{r}_t$ and $\mathbf{b}_t$ form the semantic state $\mathbf{u}_t$ defined in Section~\ref{subsec:system_overview}.

\subsubsection{Weighted-Prefix Learning Objective}
\label{subsubsec:weighted_prefix_training}
For training sequence $n$, let $L_n$ denote the number of CPIs and $y_n\in\{1,\ldots,C\}$ its ground-truth class. The belief $\mathbf{b}_{n,t}$ represents the prediction after observing the first $t$ CPIs. To align intermediate supervision with online decision epochs, supervision is applied at intervals of $K$ CPIs and begins only after at least a fraction $\rho_{\min}$ of the sequence has been observed. This excludes very short prefixes that may contain insufficient task-relevant information. The resulting supervised prefix set is
\begin{equation}
\mathcal{P}_n
=
\left\{
t\in\{K,2K,3K,\ldots\}
:
t<L_n,\quad
\frac{t}{L_n}
\geq
\rho_{\min}
\right\}.
\label{eq:valid_prefix_set}
\end{equation}
Here, $K$ is the decision interval and $\rho_{\min}$ the minimum supervised observation fraction. The complete sequence receives separate supervision. The prefix set is constructed during training, when the complete sequence length $L_n$ is known. During online operation, observations are processed as they arrive, and the model does not require $L_n$.

Each intermediate prefix is assigned a weight that increases with the observed fraction of the sequence:
\begin{equation}
w_{n,t}
=
\left(
\frac{t}{L_n}
\right)^{\kappa},
\qquad
t\in\mathcal{P}_n,
\label{eq:prefix_weight}
\end{equation}
where $\kappa$ controls the relative emphasis on later intermediate prefixes. Later prefixes generally provide more reliable semantic evidence; assigning them larger weights stabilizes training while retaining supervision at earlier decision epochs.

For classification, the loss is defined as
\begin{equation}
\ell_{\rm cls}
\left(
\mathbf{b}_{n,t},
y_n
\right)
=
-\log
\left[
\mathbf{b}_{n,t}
\right]_{y_n},
\label{eq:prefix_classification_loss}
\end{equation}
the normalized intermediate-prefix loss is
\begin{equation}
\mathcal{L}_{\rm prefix}^{(n)}
=
\frac{
1
}{
\displaystyle
\sum_{t\in\mathcal{P}_n}
w_{n,t}
}\displaystyle
\sum_{t\in\mathcal{P}_n}
w_{n,t}
\ell_{\rm cls}
\left(
\mathbf{b}_{n,t},
y_n
\right),
\label{eq:normalized_prefix_loss}
\end{equation}
and the training objective for sequence $n$ is
\begin{equation}
\mathcal{L}^{(n)}
=
\omega_{\rm f}
\ell_{\rm cls}
\left(
\mathbf{b}_{n,L_n},
y_n
\right)
+
\left(
1-\omega_{\rm f}
\right)
\mathcal{L}_{\rm prefix}^{(n)},
\label{eq:weighted_prefix_loss}
\end{equation}
where $\omega_{\rm f}$ balances full-sequence and prefix supervision.

\subsection{Semantic Utility Network}
\label{subsec:semantic_utility_network}
The semantic utility network evaluates how useful each feasible sensing profile is for the next observation block. Given the current semantic state, it predicts the utility of continuing sensing under each candidate profile. At each decision epoch, these utilities determine both continued acquisition and the profile used to perform it.

\subsubsection{Candidate-Conditioned Utility Input}
\label{subsubsec:candidate_utility_input}
At decision epoch $t$, the latent representation $\mathbf{r}_t$ summarizes accumulated task-relevant information, while the semantic belief $\mathbf{b}_t$ represents the current task-level inference. For each candidate profile $a_p\in\mathcal{A}_t^{\rm feas}$, the shared profile encoder produces its embedding
$\mathbf{g}_p=\mathcal{E}_{\theta_{\rm A}}(\mathbf{q}_p)$.
The candidate-conditioned utility input is
$\mathbf{v}_t^{(p)}=\operatorname{concat}(\mathbf{r}_t,\mathbf{b}_t,\mathbf{g}_p)$, where $\operatorname{concat}(\cdot)$ joins the three feature vectors into a single input vector.
This representation lets the utility network consider the semantic evidence and sensing characteristics of each candidate profile.

\subsubsection{Semantic Utility Prediction}
\label{subsubsec:semantic_utility_prediction}

A utility network shared across all candidate profiles predicts
\begin{equation}
U_t^{(p)}
=
\mathcal{F}_{\theta_{\rm U}}
\left(
\mathbf{v}_t^{(p)}
\right),
\qquad
a_p\in\mathcal{A}_t^{\rm feas},
\label{eq:candidate_utility_prediction}
\end{equation}
where $U_t^{(p)}\in\mathbb{R}$ represents the predicted utility of acquiring the next observation block under profile $a_p$. A larger value indicates greater expected semantic improvement after accounting for sensing cost. When sensing continues, the agent applies the highest-utility candidate to the next block.

\subsubsection{Semantic Utility Objective}
\label{subsubsec:semantic_utility_objective}
After training, the semantic belief network is fixed to construct supervised utility targets. The candidate-profile semantic gain is defined as the task-loss reduction obtained from one additional observation block:
\begin{equation}
G_{n,t}^{(p)}
=
\ell_{\rm sem}
\left(
y_n,\mathbf{b}_{n,t}
\right)
-
\ell_{\rm sem}
\left(
y_n,\mathbf{b}_{n,t'}^{(p)}
\right)
\label{eq:semantic_task_loss_reduction}
.
\end{equation}
Here, $\ell_{\rm sem}(\cdot)$ denotes the task loss. At decision epoch $t<L_n$ of training sequence $n$, $\mathbf{b}_{n,t}$ is the current belief, and $\mathbf{b}_{n,t'}^{(p)}$ is produced by the fixed belief network after processing the next observation block under $a_p\in\mathcal{A}_t^{\rm feas}$, where $t'>t$ is the next decision epoch. The additional block and ground-truth target are used only for utility-target construction, not during online sensing. For classification, we use cross-entropy,
$\ell_{\rm sem}(y,\mathbf{b})=-\log b_y$, yielding
\begin{equation}
G_{n,t}^{(p)}
=
\log
\frac{
b_{n,t',y_n}^{(p)}
}{
b_{n,t,y_n}
}.
\label{eq:semantic_log_probability_gain}
\end{equation}
A positive value indicates that the additional observation reduces task loss, whereas a negative value denotes semantic degradation. For other tasks, $\ell_{\rm sem}$ is task-specific.

Let $C_{n,t}^{(p)}$ denote the normalized sensing-resource cost of acquiring the next observation block under candidate profile $a_p$. This cost is represented by a system-dependent function:
\begin{equation}
C_{n,t}^{(p)}
=
\mathcal{C}
\left(
a_p,t'-t
\right).
\label{eq:incremental_candidate_cost}
\end{equation}
Here, $t'-t$ is the candidate-block duration, and $\mathcal{C}(\cdot)$ measures its sensing-resource cost. The supervised utility target combines this cost with the semantic gain:
\begin{equation}
U_{n,t}^{{\rm target},(p)}
=
G_{n,t}^{(p)}
-
\lambda_{\rm cost}
C_{n,t}^{(p)}.
\label{eq:semantic_utility_target}
\end{equation}
Here, $\lambda_{\rm cost}\geq0$ balances semantic gain and sensing cost.

The utility network is then trained on the decision--candidate samples using the mean-squared-error (MSE) objective:
\begin{equation}
\mathcal{L}_{\rm utility}
=
\frac{1}{|\mathcal{S}_{\rm U}|}
\sum_{(n,t,p)\in\mathcal{S}_{\rm U}}
\left(
U_{n,t}^{(p)}
-
U_{n,t}^{{\rm target},(p)}
\right)^2,
\label{eq:utility_network_loss}
\end{equation}
where $\mathcal{S}_{\rm U}$ denotes the utility-training set. With this objective, the learned utility approximates the conditional target mean given the semantic state and candidate context. Section~\ref{sec:voi_interpretation} relates this conditional mean to VoI.

\subsubsection{Profile Selection and Semantic Early Exit}
\label{subsubsec:stopping_profile_selection}

At decision epoch $t$, $U_t^{(p)}$ predicts the utility of acquiring the next block under candidate profile $a_p$. After acquisition, the semantic state and candidate utilities are updated. Their maximum across feasible profiles is defined as
\begin{equation}
U_t^{\max}
=
\max_{a_p\in\mathcal{A}_t^{\rm feas}}
U_t^{(p)}.
\label{eq:maximum_candidate_utility}
\end{equation}

When sensing continues, the agent selects
\begin{equation}
a_t^{*}
=
\underset{a_p\in\mathcal{A}_t^{\rm feas}}{\arg\max}
\;
U_t^{(p)},
\label{eq:utility_based_profile_selection}
\end{equation}
and applies $a_t^{*}$ to the next observation block. The selected profile therefore adapts the sensing-resource usage of subsequent acquisition according to the evolving semantic state.

Let $N_t^{\rm obs}$ denote the number of acquired CPIs at decision epoch $t$. To prevent premature termination, semantic early exit is enabled after $N_{\min}$ CPIs and uses the stopping rule
\begin{equation}
N_t^{\rm obs} \geq N_{\min}
\quad \text{and} \quad
U_t^{\max} \leq \tau.
\label{eq:semantic_stopping_rule}
\end{equation}
The agent then returns the current prediction; otherwise, it continues with $a_t^{*}$. Before $N_t^{\rm obs}$ reaches $N_{\min}$, the controller keeps sensing while allowing profile adaptation at each decision epoch. The parameter $\lambda_{\rm cost}$ controls the semantic--cost balance; $N_{\min}$ and $\tau$ govern early-exit eligibility.

\subsection{VoI Interpretation of the Continuation Utility}
\label{sec:voi_interpretation}

The semantic-gain target admits a VoI interpretation~\cite{kleinegesse2021sequential}. For classification, let $Y$ denote the random class label, $\mathcal{H}_t=h_t$ the observed history, and $Z_t^{(p)}$ the next block acquired under feasible profile $a_p\in\mathcal{A}_t^{\rm feas}$. We assume exact posteriors for the following VoI analysis:
\begin{equation}
\mathbf{b}_t
=
P(Y\mid\mathcal{H}_t=h_t),
\quad
\mathbf{b}_{t'}^{(p)}
=
P(Y\mid\mathcal{H}_t=h_t,Z_t^{(p)}).
\label{eq:ideal_semantic_posteriors}
\end{equation}

Let $G_t^{(p)}$ denote the population-level counterpart of the sample-wise semantic gain in~\eqref{eq:semantic_task_loss_reduction}. Under the ideal-posterior assumption, its expected log-loss reduction measures the information supplied by the next block.

\noindent\textbf{Proposition 1 (Semantic VoI of the Next Observation Block).}
Under~\eqref{eq:ideal_semantic_posteriors}, the expected log-loss reduction produced by candidate profile $a_p$ equals the conditional mutual information between the task label and the next observation block:
\begin{equation}
\mathbb{E}
\left[
G_t^{(p)}
\mid
\mathcal{H}_t=h_t
\right]
=
I
\left(
Y;
Z_t^{(p)}
\mid
\mathcal{H}_t=h_t
\right)
\geq 0.
\label{eq:voi_interpretation}
\end{equation}

\noindent\textbf{\textit{Proof.}}
Conditioned on the observed history $\mathcal{H}_t=h_t$, the expected semantic gain under profile $a_p$ is
\begin{equation}
\begin{aligned}
&\mathbb{E}\left[G_t^{(p)}\mid\mathcal{H}_t=h_t\right]
=
\mathbb{E}_{Z_t^{(p)},Y\mid h_t}
\left[
\log\frac{P(Y\mid h_t,Z_t^{(p)})}{P(Y\mid h_t)}
\right]
\\
&\quad=
\mathbb{E}_{Z_t^{(p)}\mid h_t}
\Big[
D_{\rm KL}\big(
P(Y\mid h_t,Z_t^{(p)})
\,\|\,P(Y\mid h_t)
\big)
\Big]
\\
&\quad=
I\left(Y;Z_t^{(p)}\mid\mathcal{H}_t=h_t\right).
\end{aligned}
\label{eq:voi_proof}
\end{equation}
Here, $D_{\rm KL}(\cdot\|\cdot)$ denotes the Kullback--Leibler (KL) divergence. Since the KL divergence is nonnegative, the expected semantic gain is also nonnegative, although an individual realized gain may be negative. This proves the result. \hfill$\square$

Let $C_t^{(p)}$ denote the corresponding candidate-block cost in the population formulation. For candidate profile $a_p$, this formulation defines the ideal cost-adjusted VoI as
\begin{equation}
V_t^{(p)}
=
I
\left(
Y;
Z_t^{(p)}
\mid
\mathcal{H}_t=h_t
\right)
-
\lambda_{\rm cost}C_t^{(p)}.
\label{eq:population_continuation_value}
\end{equation}
Thus, under the ideal-posterior assumption, $V_t^{(p)}>0$ when the next block provides more expected task information than the sensing-resource penalty. This relation provides a theoretical basis for the learned utility.

\noindent\textbf{Remark 1 (VoI Interpretation Versus Learned Utility).}
Under the MSE objective in~\eqref{eq:utility_network_loss}, the population-optimal predictor for the candidate-conditioned input $\mathbf{v}_t^{(p)}$ is
\begin{equation}
U^{*}
\left(
\mathbf{v}_t^{(p)}
\right)
=
\mathbb{E}
\left[
U_t^{{\rm target},(p)}
\mid
\mathbf{v}_t^{(p)}
\right].
\label{eq:mse_conditional_utility}
\end{equation}
Here, $U_t^{{\rm target},(p)}$ is the population counterpart of the supervised target in~\eqref{eq:semantic_utility_target}, constructed using the fixed belief network. The learned utility therefore estimates the expected task-loss reduction minus sensing cost, conditioned on the semantic state and candidate context. Its interpretation in terms of conditional mutual information holds under the ideal-posterior assumption above. Ground-truth labels and future candidate observations are used only for utility-target construction during training and are not included in the online utility input.

\section{System Design of Agentic SemS}
\label{sec:system_design_implementation}
This section presents a reference implementation of Agentic SemS, covering its network architectures, training, and online operation. Section V evaluates this implementation on a specific sensing task using multiple sensing-profile configurations.

\subsection{Semantic Belief Network Implementation}
\label{subsec:semantic_belief_sjoetwork_settings}

\subsubsection{CPI-Level Feature Extraction}
At CPI $t$, the reference sensing front end receives the measurement tuple
\begin{equation}
\mathbf{m}_t
=
\left(
\mathbf{C}_t,
\mathbf{Q}_t
\right),
\label{eq:belief_measurement_input}
\end{equation}
where $\mathbf{C}_t=\{\mathbf{C}_{t,n}\}_{n=1}^{N_{{\rm pkt},t}}$ is the CSI packet sequence, and each $\mathbf{C}_{t,n}\in\mathbb{C}^{N_{{\rm sc},t}\times M_t}$ contains measurements over $N_{{\rm sc},t}$ retained subcarriers and $M_t$ receive antennas. CSI is widely used in wireless sensing because it preserves fine-grained motion-induced channel variations across subcarriers and antennas, supporting delay--Doppler feature extraction~\cite{wu2025ISAC}. The compact-radio-measurement sequence is $\mathbf{Q}_t=\{\mathbf{d}_{t,\ell}\}_{\ell=1}^{L_t}$, where $\mathbf{d}_{t,\ell}\in\mathbb{R}^{N_{{\rm W},t}}$ contains $N_{{\rm W},t}$ radio indicators, such as received-power, link-quality, and interference measurements. The active profile determines which of these inputs are available.

For CSI extraction, we follow the phase-compensation and feature-construction procedures in~\cite{wang2025siso}. Phase compensation mitigates timing- and carrier-frequency-offset variations, while temporal-mean removal suppresses static components. Adaptive spectral processing combines 32 range bins from a range-domain fast Fourier transform (FFT) with 128 Doppler bins from a temporal FFT, forming a range--Doppler spectrum over the intervals supported by the sensing configuration. Compact radio measurements are processed through an appropriate statistical or spectral front end to extract task-relevant temporal variations~\cite{wang2026rethinking}. In the reference implementation, the selected compact-radio sequence is mean-centered to suppress static components, and a 128-point FFT extracts its CPI-level temporal spectral feature. Both front ends produce a 128-dimensional CPI-level spectral feature. This feature is augmented with seven spectrum-derived statistics: log standard deviation, log total energy, log peak magnitude, negative- and positive-frequency energy ratios, normalized spectral centroid, and normalized spectral width, resulting in a 135-dimensional CPI observation. The reference encoder in~\eqref{eq:observation_encoding} takes the form
\begin{equation}
\mathcal{E}_{\theta_{\rm o}}(\mathbf{o}_t)
=
\mathbf{W}_{\rm o}\operatorname{LN}(\mathbf{o}_t)
+
\boldsymbol{\beta}_{\rm o}
,
\label{eq:reference_observation_encoder}
\end{equation}
where $\operatorname{LN}(\cdot)$ denotes layer normalization, $\mathbf{W}_{\rm o}\in\mathbb{R}^{192\times135}$ is the learnable projection matrix, and $\boldsymbol{\beta}_{\rm o}\in\mathbb{R}^{192}$ is the learnable bias vector. The encoder projects the 135-dimensional observation to a 192-dimensional embedding.

\subsubsection{Profile-Context Construction}
\label{subsubsec:implementation_profile_context}
Each sensing profile is represented by a fixed-dimensional context
\begin{equation}
\mathbf{q}_p
=
\left[
\boldsymbol{\xi}_p^{\mathsf T},
c_p,
\boldsymbol{\iota}_p^{\mathsf T}
\right]^{\mathsf T}
\in\mathbb{R}^{N_{\rm q}}
.
\label{eq:reference_profile_context}
\end{equation}
Here, $\boldsymbol{\xi}_p\in\mathbb{R}^{N_{\xi}}$ contains numerical sensing-configuration attributes, such as the sampling rate, signal-to-noise ratio (SNR), subcarrier usage, and temporal configuration. Before encoding, continuous attributes are scaled to comparable numerical ranges because they may have different physical units and magnitudes. The scalar $c_p\in[0,1]$ is a normalized per-CPI sensing-resource score associated with profile $a_p$. The binary vector $\boldsymbol{\iota}_p\in\{0,1\}^{N_{\iota}}$ indicates the available measurement branches and applicable configuration attributes of profile $a_p$.
The attribute set is determined by the sensing interface and remains fixed across profiles within a deployment. A two-layer profile encoder with a 64-dimensional hidden layer maps $\mathbf{q}_p$ to a 192-dimensional embedding shared by the belief and utility networks. After joint training with the belief network, the encoder is frozen and reused by the utility network.

\subsubsection{Network Architecture}

The 192-dimensional observation, active-profile, and positional embeddings form $\mathbf{x}_t$ in~\eqref{eq:profile_conditioned_cpi_input}. The causal Transformer $\mathcal{F}_{\theta_{\rm B}}$ uses four self-attention layers with six heads, a 192-dimensional model, and a 768-dimensional feed-forward layer. At each CPI $t$, it outputs $\mathbf{r}_t\in\mathbb{R}^{192}$, and the classification head maps $\mathbf{r}_t$ to the $C$-dimensional semantic belief $\mathbf{b}_t\in\mathbb{R}^{C}$.

\subsubsection{Two-Stage Weighted-Prefix Training}
The weighted-prefix objective is evaluated at intermediate prefixes separated by $K=4$ CPIs. In this implementation, consecutive CPIs use overlapping windows with a 32-ms stride; hence, $K=4$ places decision epochs 128~ms apart. Prefix supervision begins after 20\% of a sequence is observed, and the prefix weights use $w_{n,t}=(t/L_n)^{\kappa}$ with $\kappa=2$. Training proceeds in two stages. Stage~1 uses fixed-profile trajectories to learn reliable representations at intermediate prefixes and complete sequences, with $\omega_{\rm f}=0.7$. Stage~2 improves robustness to online profile changes using $\omega_{\rm f}=0.5$, with 30\% fixed-profile and 70\% mixed-profile trajectories generated using feasible adjacent profile transitions between observation blocks. Stage~1 trains for 100 epochs with a learning rate of $10^{-3}$, and Stage~2 trains for 50 epochs with a learning rate of $10^{-4}$. Both stages use batch size 128, AdamW with weight decay $10^{-4}$, cosine annealing to $10^{-6}$, gradient clipping with a maximum norm of 1, and random seed 42.

\subsubsection{KV-Cached Inference}
For the causal Transformer, the previously computed key and value tensors are retained as
\begin{equation}
\mathcal{C}_t^{\rm KV}
=
\left\{
\mathbf{K}_{1:t}^{(\ell)},
\mathbf{V}_{1:t}^{(\ell)}
\right\}_{\ell=1}^{L_{\rm B}},
\label{eq:semantic_kv_cache}
\end{equation}
where $L_{\rm B}$ is the number of Transformer layers. A newly acquired CPI is processed incrementally according to
\begin{equation}
\left(
\mathbf{r}_t,
\mathcal{C}_t^{\rm KV}
\right)
=
\mathcal{F}_{\theta_{\rm B}}
\left(
\mathbf{x}_t;
\mathcal{C}_{t-1}^{\rm KV}
\right).
\label{eq:kv_cached_semantic_belief_update}
\end{equation}
In the reference implementation, streaming inference uses a KV cache for at most 128 CPI-level tokens, with each token corresponding to one acquired CPI input $\mathbf{x}_t$. The cache capacity is an implementation parameter and can be increased for applications requiring a longer sensing horizon, subject to available memory. The cache also persists across profile transitions because the stored tokens retain their profile embeddings.

\subsection{Semantic Utility Network and Online Agent Implementation}
\label{subsec:semantic_utility_network_settings}

\subsubsection{Candidate-Conditioned Utility Input}
With the semantic belief network fixed, the utility network concatenates the 192-dimensional latent representation $\mathbf{r}_t$, $C$-dimensional semantic belief $\mathbf{b}_t$, and 192-dimensional candidate-profile embedding $\mathbf{g}_p$ for each $a_p\in\mathcal{A}_t^{\rm feas}$ at every decision epoch. This forms the $(384+C)$-dimensional candidate-conditioned input $\mathbf{v}_t^{(p)}$.

\subsubsection{Utility Network Architecture}
The candidate-conditioned input is processed by a lightweight multilayer perceptron with hidden dimensions of 256 and 128. The first linear layer is followed by LayerNorm and a Gaussian error linear unit (GELU)~\cite{hendrycks2016gelu}, while the second hidden layer also uses GELU. A final linear layer produces a continuation utility. The same network is shared across all candidate profiles.

\subsubsection{Utility-Target Generation}
Utility targets are generated at decision epochs spaced by $K=4$ CPIs. At each epoch, a candidate is evaluated using the next four-CPI observation block, and a training sample is constructed only when this complete block is available. For each feasible candidate, the target follows~\eqref{eq:semantic_log_probability_gain} and~\eqref{eq:semantic_utility_target}. Here, $c_p$ denotes the normalized per-CPI sensing-resource cost, so the candidate block has cost $C_{n,t}^{(p)}=4c_p$. Because this factor of four is fixed and common to all candidates, the reported implementation-level $\lambda_{\rm cost}$ absorbs this factor, and the utility target uses $c_p$ directly. Cumulative sensing cost remains calculated at the CPI level. Because $\lambda_{\rm cost}$ is part of the supervised target, each value requires regenerated targets and a separately trained utility checkpoint. The parameters $N_{\min}$ and $\tau$ are adjusted only at inference.

\begin{table*}[t]
\centering
\caption{Ordered sensing profiles and normalized per-CPI sensing-resource costs under the reference weighting.}
\label{tab:implementation_profiles}
\fontsize{8}{10}\selectfont
\renewcommand{\arraystretch}{1.10}
\setlength{\tabcolsep}{5pt}
\begin{tabular}{c|c|c|c|c|c|c|c|c}
\Xhline{1pt}
\multirow{2}{*}{Profile}
& Feature
& Subcarriers
& Sampling rate
& Effective SNR
& CPI length
& CPI stride
& CPI budget
& Reference cost
\\
\cline{2-9}
& $\psi$
& $N_{\rm sc}$
& $F_s$ (kHz)
& $\Delta\gamma$ (dB)
& $W_{\rm CPI}$
& $I_{\rm CPI}$
& $L_{\rm seq}$
& $c_p(\mathbf{w}_{\rm ref})$
\\
\Xhline{1pt}
Lightweight & RSSI & -- & 0.25 & 20 & 32 & 8 & 64 & 0.236 \\
\hline
Low & CSI & 10 & 0.25 & 25 & 32 & 8 & 48 & 0.528 \\
\hline
Low-Medium & CSI & 15 & 0.5 & 25 & 64 & 16 & 64 & 0.655 \\
\hline
Medium & CSI & 20 & 0.5 & 30 & 64 & 16 & 96 & 0.720 \\
\hline
Medium-High & CSI & 25 & 1 & 30 & 128 & 32 & 96 & 0.936 \\
\hline
High & CSI & 30 & 1 & -- & 128 & 32 & 128 & 1.000 \\
\Xhline{1pt}
\end{tabular}
\vspace{-1.5mm}
\end{table*}

\begin{figure*}[t]
\centering
\includegraphics[width=0.16\textwidth]{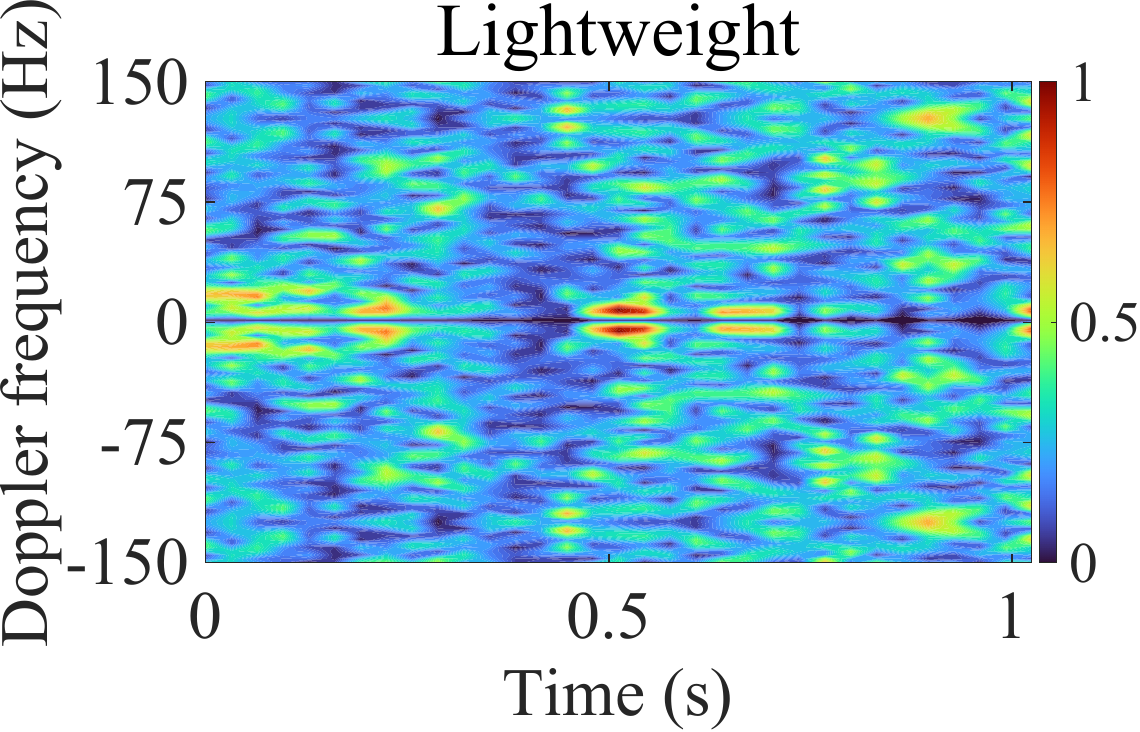}
\includegraphics[width=0.16\textwidth]{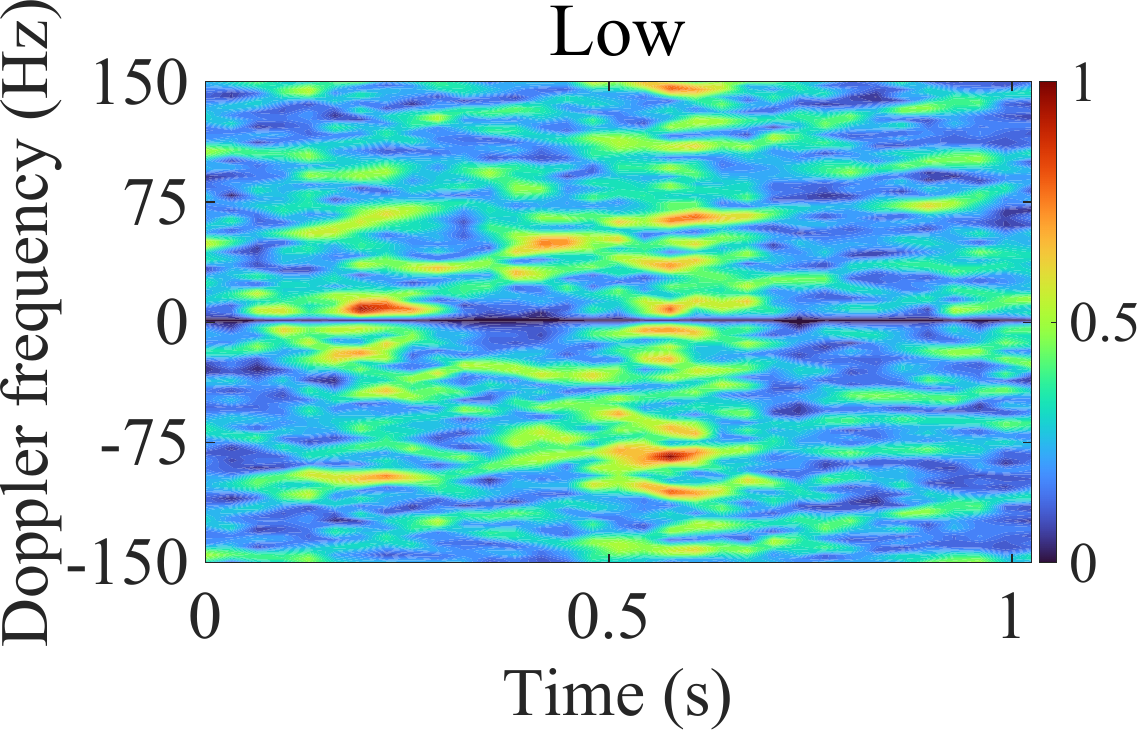}
\includegraphics[width=0.16\textwidth]{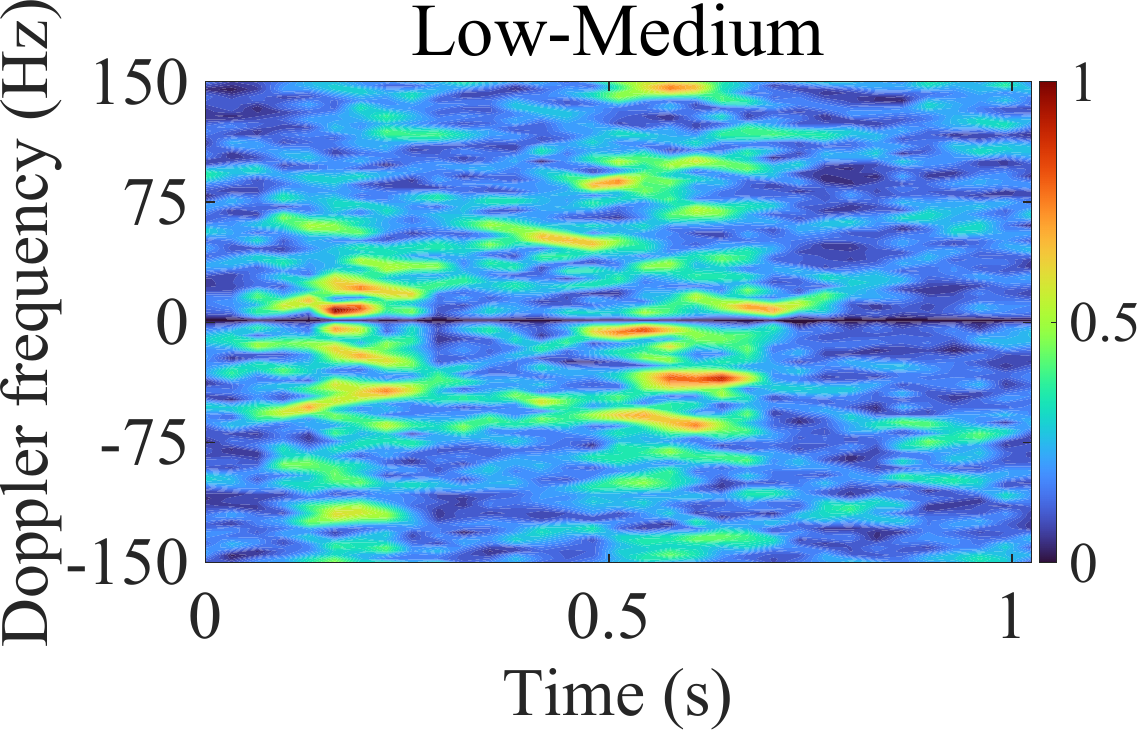}
\includegraphics[width=0.16\textwidth]{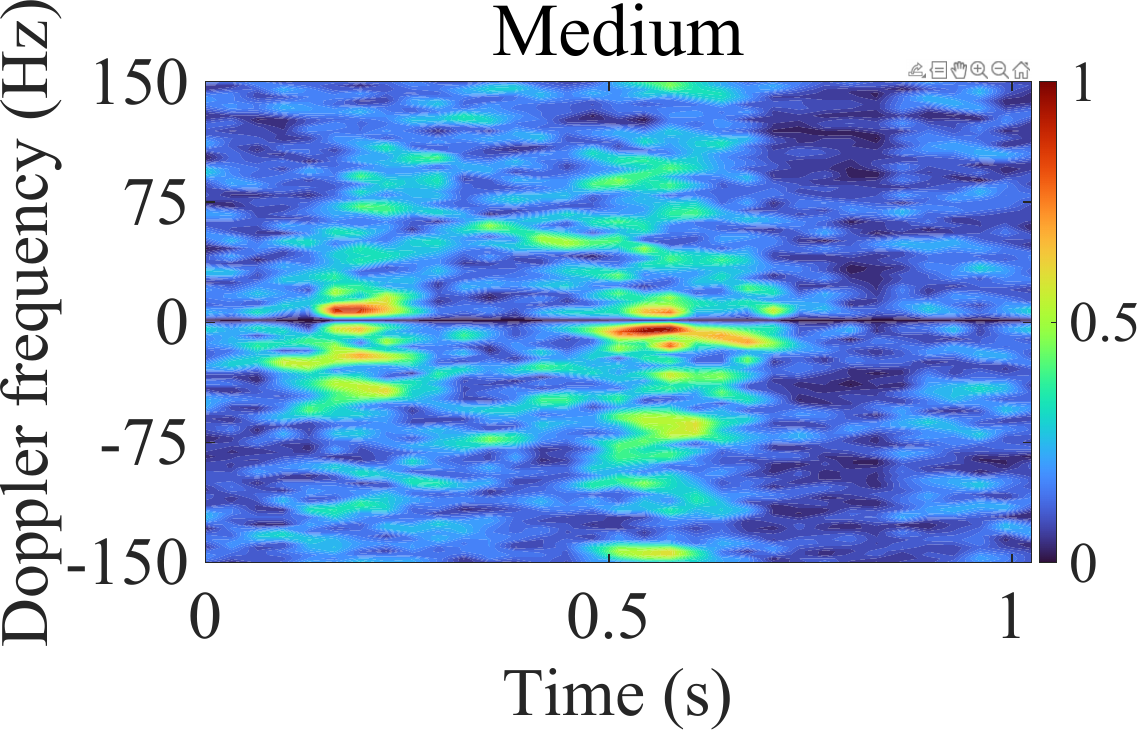}
\includegraphics[width=0.16\textwidth]{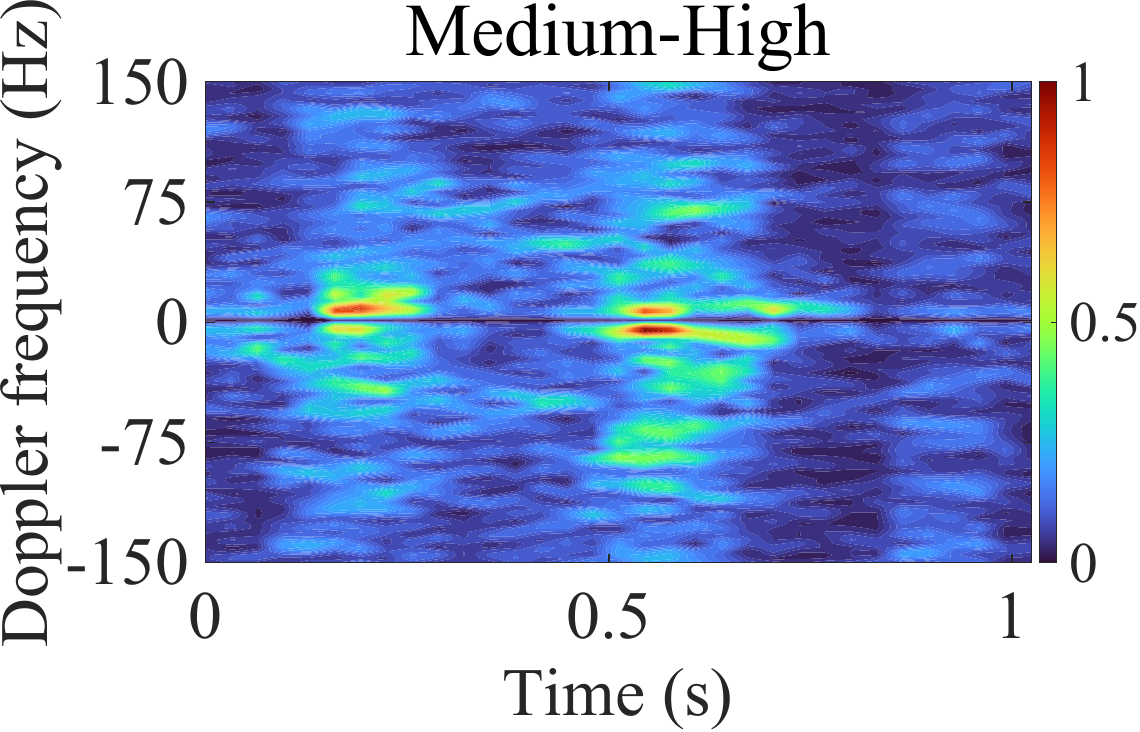}
\includegraphics[width=0.16\textwidth]{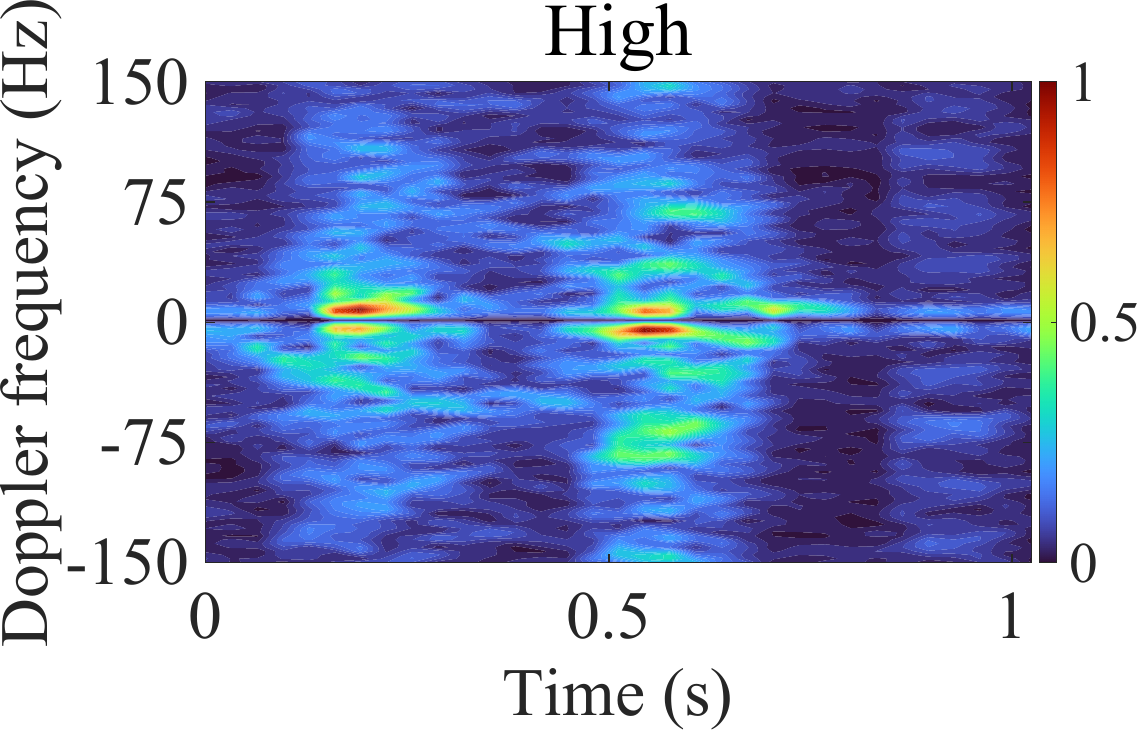}
\caption{Micro-Doppler spectrograms under the six sensing profiles, ordered from Lightweight to High from left to right.}
\label{fig:profile_time_doppler}
\vspace{-1.5em}
\end{figure*}

\subsubsection{Utility Network Training}
The utility network is trained for 30 epochs using record-level sampling. Each epoch samples 2,000 training records; for each record, all decision epochs with a complete next observation block and their feasible candidates contribute candidate-level training examples. Training uses the MSE objective in~\eqref{eq:utility_network_loss} and AdamW with a learning rate of $10^{-3}$, weight decay of $10^{-4}$, cosine annealing to $10^{-5}$, an accumulation size of 128 records, and random seed 42. For each $\lambda_{\rm cost}$, validation retains the minimum-MSE checkpoint.

\subsubsection{Online Profile Adaptation and Semantic Early Exit}
\label{subsubsec:online_profile_adaptation_implementation}
The online agent maintains the active profile, current semantic state, and Transformer KV cache as persistent runtime states. In the reference controller, the sensing profiles are ordered by resource usage and the transition rule in~\eqref{eq:feasible_profile_set} is instantiated as
\begin{equation}
\mathcal{A}_t^{\rm feas}
=
\left\{
a_p\in\mathcal{A}^{\rm com}:|p-s_t|\leq 1
\right\},
\label{eq:implementation_adjacent_profile_set}
\end{equation}
which permits the current profile and its available adjacent lower- and higher-resource profiles. Stay retains the current profile, whereas Down and Up select the adjacent profiles. This transition rule is an implementation choice, not a requirement of Agentic SemS. The controller uses only acquired CPI history, active-profile context, and feasible candidates; it receives neither the complete record length nor future observations. The belief network processes each acquired CPI incrementally, while the utility network evaluates continuation profiles after each observation block is acquired. If sensing continues, the highest-utility candidate is selected for the next block. After acquisition, the semantic state and candidate utilities are updated. The KV cache is retained across profile transitions and reuses the observed context. Profile adaptation is available from the first decision epoch, whereas semantic early exit becomes eligible after $N_{\min}$ has been reached. Thereafter, sensing continues under the selected profile when the maximum candidate utility exceeds $\tau$; otherwise, it stops and returns the current task prediction as the semantic output. Evaluation also terminates when a complete next block is unavailable or the selected profile reaches its configured CPI budget; these limits are enforced by the sensing environment rather than provided to the agent as future information.

\section{Experimental Evaluation}
\label{sec:experimental_results}
We evaluate the reference implementation on a six-class WiFi gesture-recognition task. The dataset, sensing-profile realizations, controller configuration, and evaluation protocol are described first, followed by the experimental results.

\subsection{Experimental Setup}
\label{sec:experimental_setup}

\subsubsection{Dataset and Partitioning}
\label{subsec:experimental_dataset}
We use Widar3.0~\cite{zhang2021widar3} for six-class WiFi gesture recognition. It was collected in the 5-GHz band with Intel 5300 network interface cards, one transmitter, and multiple receivers. Each receiver provides CSI over 30 subcarriers and three receive antennas, together with RSSI at a 1-kHz packet rate. It contains 165,724 records of six gestures: Clap, Draw-O, Draw-Zigzag, Push \& Pull, Slide, and Sweep, from 16 subjects across three environments, five locations, and five orientations. Each record contains one pre-segmented gesture sequence under diverse motion and propagation conditions. A class-stratified 70/10/20 split yields 116,006 training, 16,572 validation, and 33,146 test records with comparable proportions. All six profile realizations of each source record remain in the same partition to prevent cross-profile leakage. The training split fits both networks, validation selects their checkpoints and controller parameters $(\lambda_{\rm cost}, N_{\min}, \tau)$, and testing is used only for final evaluation.

\begin{table*}[t]
\centering
\caption{Recognition and resource use on the test set for fixed and adaptive sensing controls.}
\label{tab:fixed_cpi_main_comparison}
\fontsize{8}{10}\selectfont
\renewcommand{\arraystretch}{1.12}
\setlength{\tabcolsep}{9pt}
\begin{tabular}{l|c|c|c|c|c|c|c}
\Xhline{1pt}
Method & Profile policy & $\lambda_{\rm cost}$ & $\tau$ & Macro-F1 (\%) & $\overline{P}$ & $\overline{R}_{\rm obs}$ (\%) & $\overline{C}$ \\
\Xhline{1pt}
Full-sequence High & Fixed High & -- & Disabled & \textbf{90.70} & 1.000 & 100.00 & 1.000 \\
\hline
Fixed High with early exit & Fixed High & 0.005 & $-0.05$ & 90.16 & 1.000 & 88.51 & 0.885 \\
\hline
Full-sequence Medium-High & Fixed Medium-High & -- & Disabled & 87.62 & 0.936 & 100.00 & 0.936 \\
\hline
Fixed Medium-High with early exit & Fixed Medium-High & 0.005 & $-0.025$ & 86.80 & 0.936 & 86.90 & 0.813 \\
\hline
Adaptive profiles without early exit & Adaptive & 0.005 & Disabled & 86.76 & 0.852 & 100.00 & 0.852 \\
\hline
Agentic SemS reference & Adaptive & 0.005 & $0$ & 85.79 & 0.866 & 86.74 & 0.747 \\
\hline
Agentic SemS high-F1 & Adaptive & 0.005 & $-0.20$ & \textbf{86.75} & 0.852 & 99.87 & 0.851 \\
\Xhline{1pt}
\end{tabular}
\vspace{-.5em}
\end{table*}

\subsubsection{Emulated Sensing Profiles}
\label{subsec:emulated_sensing_profiles}
We construct six ordered sensing profiles from a prescribed set feasible under the communication constraints, as summarized in Table~\ref{tab:implementation_profiles}. For the WiFi realization evaluated here, $\psi\in\{\mathrm{CSI},\mathrm{RSSI}\}$ identifies the sensing feature type. The Lightweight profile uses the RSSI field provided by Widar3.0 as its compact received-power measurement, whereas the other profiles use CSI to preserve fine-grained subcarrier responses. For the CSI-based profiles, the range--Doppler spectrum is averaged over the 32 retained range bins to obtain a 128-dimensional micro-Doppler vector. The five CSI-based profiles retain $N_{\rm sc}\in\{10,15,20,25,30\}$ subcarriers. Across the six profiles, sampling rates $F_s\in\{0.25,0.5,1\}$~kHz are obtained by downsampling the original 1-kHz Widar3.0 measurement stream, and effective signal-to-noise ratio (SNR) settings $\Delta\gamma\in\{20,25,30\}$~dB are applied where applicable. Larger configurations preserve richer sensing information at greater measurement and processing demand; the High profile retains the original observation. Each Widar3.0 record contributes one receive-antenna stream shared across all six profile realizations.

CPI lengths $W_{\rm CPI}\in\{32,64,128\}$ packets use quarter-length strides $I_{\rm CPI}\in\{8,16,32\}$ packets. Together with the profile-dependent sampling rate, each CPI spans 128~ms and produces one CPI-level observation every 32~ms across all profiles. The sequence length $L_{\rm seq}\in\{48,64,96,128\}$ specifies the maximum CPI budget configured for each emulated profile. Because Widar3.0 provides pre-segmented gesture records, each profile realization is also bounded by the corresponding source-record duration. For each source recording, all profile observations are generated offline and temporally aligned over their common interval so that the same CPI represents the same underlying time interval, enabling candidate-specific utility targets to be constructed from the subsequent observation block. During online evaluation, the controller processes only observations from its selected profiles and updates the semantic state sequentially until semantic early exit or an environment-imposed availability limit is reached. Fig.~\ref{fig:profile_time_doppler} compares the resulting micro-Doppler spectrograms; higher profiles show clearer structure and richer information.

\subsubsection{Normalized Sensing-Resource Cost}
\label{subsubsec:normalized_sensing_cost}
We quantify the relative sensing-resource requirements of profile $a_p$ using
\begin{equation}
\boldsymbol{\phi}_p
=
\left[
\frac{N_{{\rm sc},p}}{30},
\frac{F_{s,p}}{1~\mathrm{kHz}},
\widetilde{\gamma}_p,
\frac{W_{{\rm CPI},p}}{128},
\eta_{\psi,p}
\right]^{\mathsf T},
\label{eq:profile_resource_vector}
\end{equation}
where the components represent normalized subcarrier usage, sampling rate, observation quality, CPI length, and feature representation. For $\Delta\gamma_p\in\{20,25,30\}$~dB, $\widetilde{\gamma}_p=\Delta\gamma_p/(35~\mathrm{dB})$, while $\widetilde{\gamma}_p=1$ for the undegraded High profile. The representation component is $\eta_{\psi,p}=1$ for CSI and $0.2$ for RSSI. The 35-dB denominator and the factor 0.2 are fixed reference scales used consistently across all experiments; they define relative profile costs rather than calibrated energy ratios. The RSSI profile has zero subcarrier usage, and CPI stride is omitted because it is determined by CPI length.

For nonnegative resource weights $\mathbf{w}=[w_1,\ldots,w_5]^{\mathsf T}$ satisfying $\sum_{j=1}^{5}w_j=1$, the normalized sensing-resource cost is
$c_p(\mathbf{w})=\mathbf{w}^{\mathsf T}\boldsymbol{\phi}_p$.
The weights are fixed rather than learned. The reference weights $\mathbf{w}_{\rm ref}$ are obtained by normalizing $[1,1,0.75,0.5,1]^{\mathsf T}$. This setting gives comparable importance to subcarrier usage, sampling rate, and feature representation, with smaller contributions from observation quality and CPI length. The resulting $c_p$ is a relative resource index rather than a measurement of physical energy consumption. Table~\ref{tab:implementation_profiles} reports $c_p(\mathbf{w}_{\rm ref})$ for each profile, with unit cost for High. Cumulative sensing cost sums these per-CPI costs over the acquired sequence, using the same weights in all experiments. For this WiFi realization, the context in~\eqref{eq:reference_profile_context} has 11 entries: the six configuration attributes in Table~\ref{tab:implementation_profiles}, the per-CPI cost $c_p$, and four binary applicability indicators for compact measurements, CSI, subcarrier count, and SNR.

\subsubsection{Controller Parameter Selection}
\label{subsubsec:controller_parameter_selection}
Controller parameters are selected exclusively on the validation set. We evaluate six utility checkpoints corresponding to $\lambda_{\rm cost}\in\{0.005,0.01,0.025,0.05,0.075,0.10\}$ and nine $\tau$ values from $-0.20$ to $0$ in increments of $0.025$. Based on the validation analysis, $N_{\min}=8$ CPIs is used for all exit-enabled settings. Validation selects $(\lambda_{\rm cost},\tau)=(0.005,-0.20)$ for the high-F1 setting and $(0.005,0)$ for the resource-efficient reference setting. All parameters are fixed before test evaluation. Adaptive variants use the same randomly sampled initial profile for each record, generated with seed 42.

\begin{figure}[t]
\centering
\scalebox{1}[1]{\includegraphics[width=0.9\columnwidth]{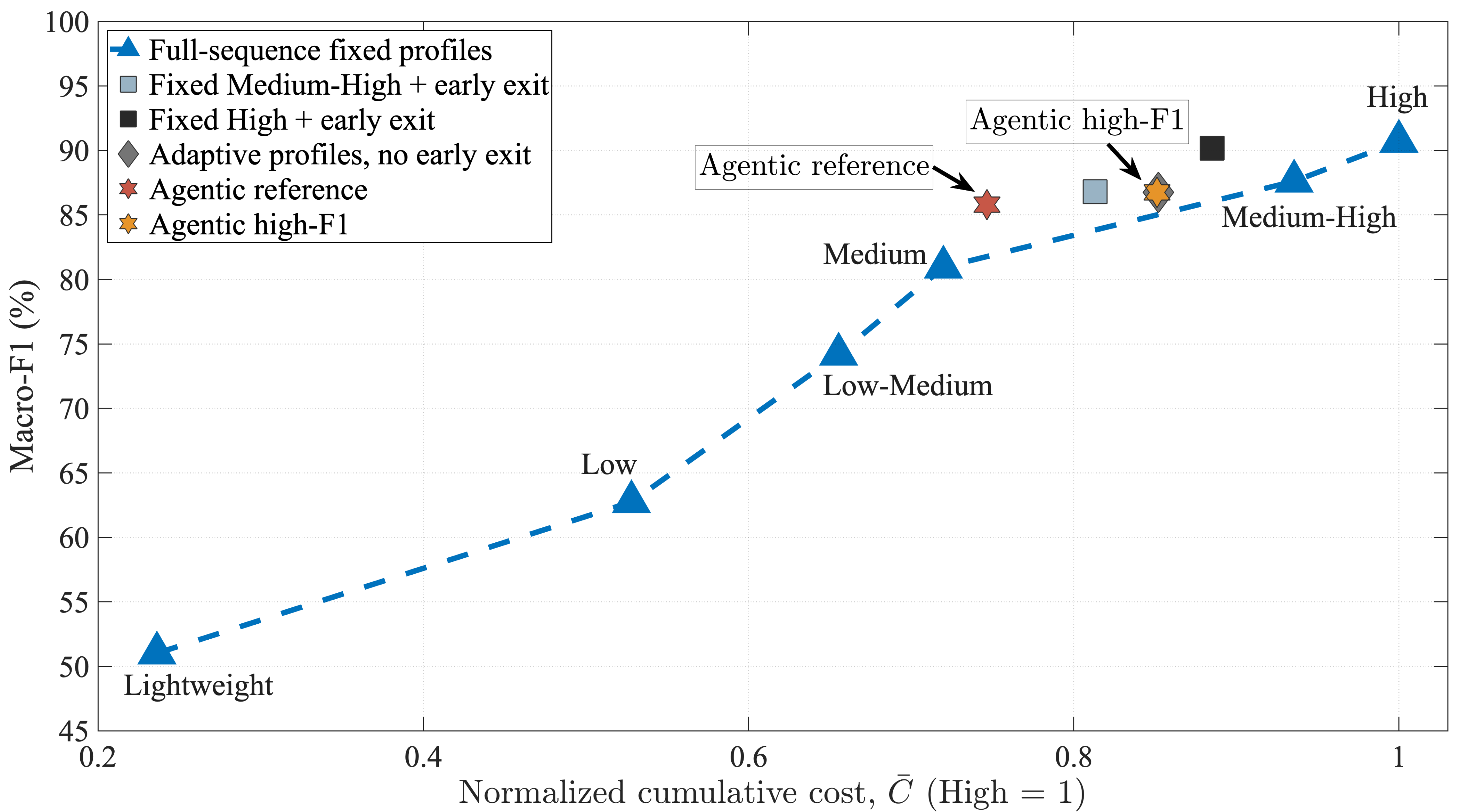}}
\caption{Macro-F1 versus sensing cost across fixed and adaptive strategies.}
\label{fig:accuracy_resource_tradeoff}
\vspace{-1.5em}
\end{figure}

\subsection{Evaluation Metrics}
\label{subsec:implementation_evaluation_metrics}

Recognition performance is evaluated using Macro-F1. For record $n$, let $t_n^{\rm stop}$ denote the number of acquired CPIs, $L_n$ the number of valid CPIs in the complete record, $s_{n,t}$ the profile active at CPI $t$, and $c_{\rm H}$ the per-CPI cost of High. Unlike the profile-level budget $L_{\rm seq}$, $L_n$ is determined by the available duration of record $n$ and is used only for metric normalization, not as a controller input. We define the normalized average profile cost and observation ratio as
\begin{equation}
P_n
=
\frac{1}{t_n^{\rm stop}}
\sum_{t=1}^{t_n^{\rm stop}}
\frac{c_{s_{n,t}}}{c_{\rm H}},
\qquad
R_n
=
\frac{t_n^{\rm stop}}{L_n},
\label{eq:implementation_profile_observation_metrics}
\end{equation}
and the normalized cumulative sensing cost as
\begin{equation}
C_n
=
\frac{
1
}{
L_n c_{\rm H}
}\displaystyle
\sum_{t=1}^{t_n^{\rm stop}}
c_{s_{n,t}}
=
P_n R_n.
\label{eq:implementation_cost_metrics}
\end{equation}
Here, $P_n$ characterizes the average sensing-profile cost during acquisition, $R_n$ measures the fraction of the complete sequence that is observed, and $C_n$ captures their combined effect on cumulative sensing-resource use.

For a test set of $N$ records, we report
$\overline{P}=\frac{1}{N}\sum_{n=1}^{N}P_n$,
$\overline{R}_{\rm obs}=\frac{1}{N}\sum_{n=1}^{N}R_n$,
$\overline{C}=\frac{1}{N}\sum_{n=1}^{N}C_n$, and
$\overline{L}_{\rm obs}=\frac{1}{N}\sum_{n=1}^{N}t_n^{\rm stop}$.
In comparison with full-sequence High, $1-\overline{P}$, $1-\overline{R}_{\rm obs}$, and $1-\overline{C}$ quantify reductions in average profile cost, observation duration, and cumulative sensing cost, respectively. These metrics distinguish the effects of profile adaptation and semantic early exit while also capturing their combined resource impact. We additionally report the semantic early-exit rate and the proportions of continuation decisions that retain, increase, or decrease the sensing profile.

\subsection{Evaluation Protocol and Baselines}
\label{subsec:evaluation_protocol_baselines}
The evaluation separates adaptive profile selection from semantic early exit. Full-sequence High is the common reference, using the highest-resource configuration without early termination. Fixed High with semantic early exit varies only observation duration, whereas adaptive profiles without early exit vary only the profile sequence. The complete Agentic SemS controller combines both mechanisms; full-sequence Medium-High and its early-exit variant provide lower-resource fixed-profile references. All exit-enabled methods use the same $N_{\min}=8$-CPI minimum-observation setting without access to the future record boundary. The threshold $\tau$ is selected on validation data and fixed before test evaluation. For each fixed-profile early-exit baseline, the continuation set contains only the selected fixed profile, and its predicted continuation utility is compared with $\tau$. Higher-F1 and Lower-cost points characterize the resulting duration-control tradeoffs.

\subsection{Overall Recognition--Resource Performance}
\label{subsec:resource_control_decomposition}

\subsubsection{Fixed-Profile Recognition and Cost}
Fig.~\ref{fig:accuracy_resource_tradeoff} shows the six full-sequence fixed-profile baselines evaluated on the test set as blue triangular markers connected by the dashed curve. All baselines use the same final causal Transformer checkpoint and differ only by profile. From Lightweight to High, Macro-F1 increases from 51.02\% to 62.75\%, 74.19\%, 80.95\%, 87.62\%, and 90.70\%, respectively. The corresponding normalized cumulative costs are
$\overline{C}=\{0.236,0.528,0.655,0.720,0.936,1.000\}$.
Since each baseline uses a single profile throughout the complete sequence, its normalized cumulative cost is identical to the corresponding normalized per-CPI cost in Table~\ref{tab:implementation_profiles}. This recognition--cost hierarchy provides the controller with multiple operating points for adapting sensing-resource use to the evolving semantic state.

\begin{figure}[t]
\centering
\scalebox{1}[1]{
\includegraphics[width=.85\columnwidth]
{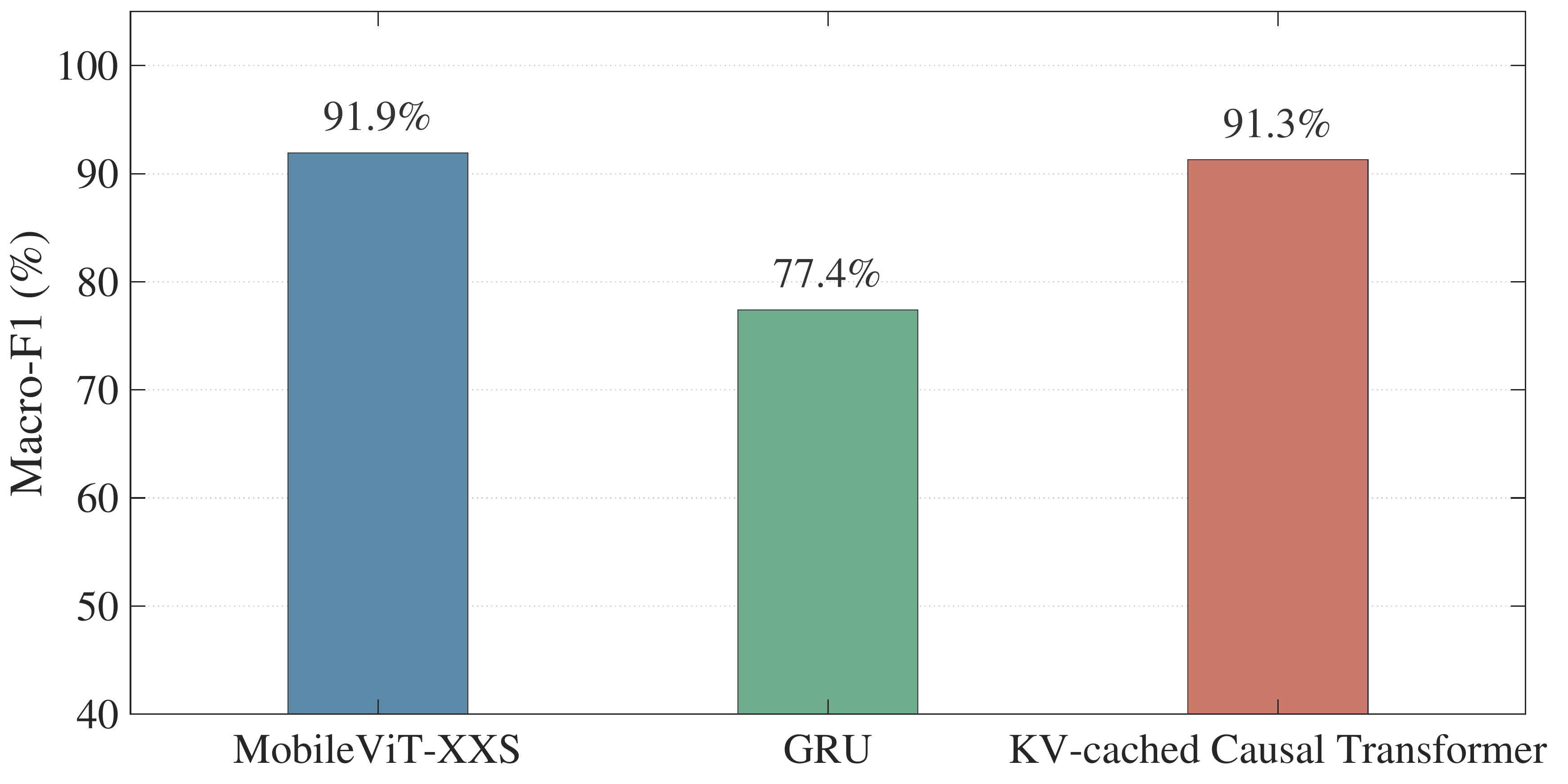}}
\caption{High-profile Macro-F1 comparison across belief networks.}
\label{fig:belief_network_comparison}
\vspace{-1.5em}
\end{figure}

\begin{figure*}[t]
\centering
\scalebox{1}[0.92]{\includegraphics[width=0.3\textwidth]{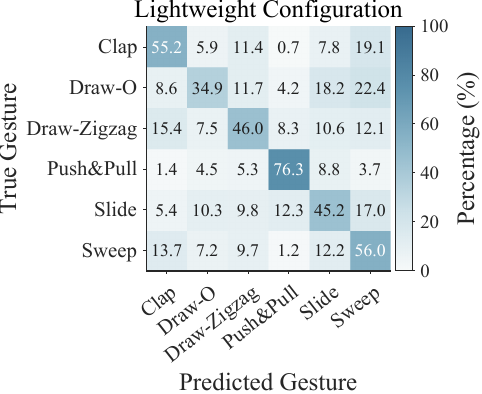}}
\scalebox{1}[0.92]{\includegraphics[width=0.3\textwidth]{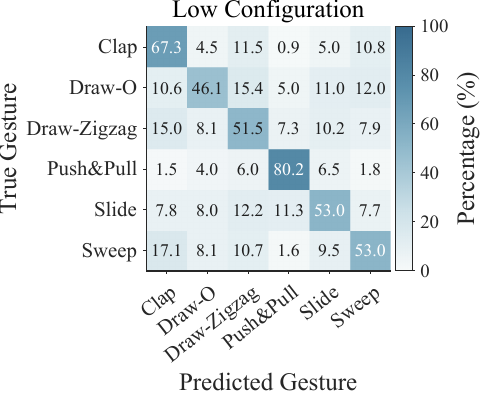}}
\scalebox{1}[0.92]{\includegraphics[width=0.3\textwidth]{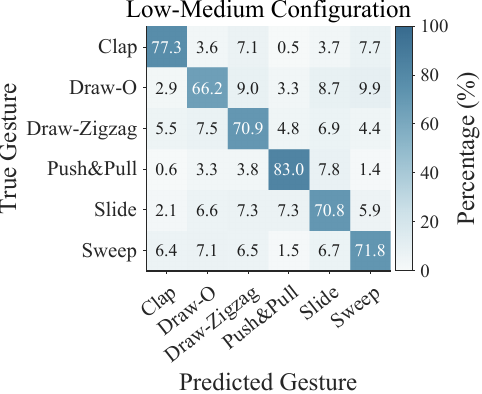}}
\\[1em]
\scalebox{1}[0.92]{\includegraphics[width=0.3\textwidth]{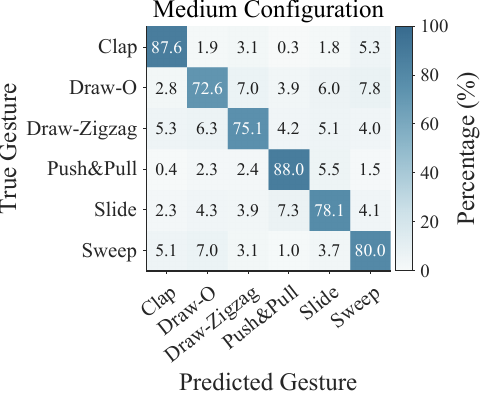}}
\scalebox{1}[0.92]{\includegraphics[width=0.3\textwidth]{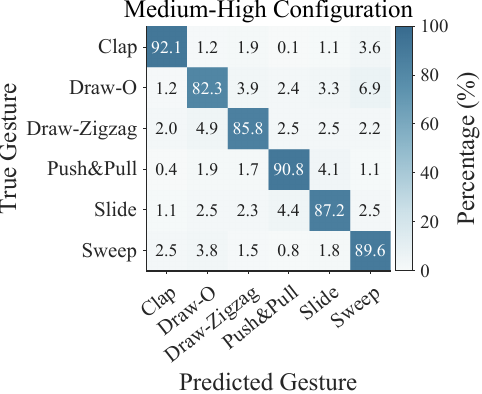}}
\scalebox{1}[0.92]{\includegraphics[width=0.3\textwidth]{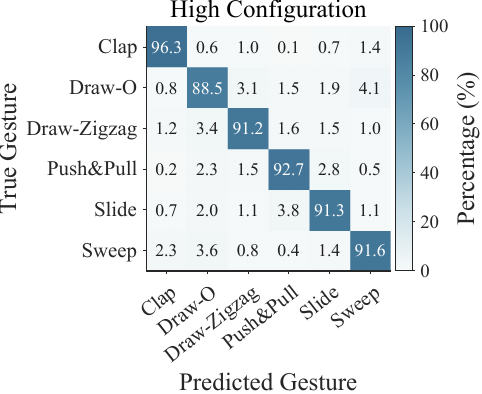}}
\caption{Class-normalized confusion matrices of the complete-sequence MobileViT-XXS models across the six emulated sensing profiles.}
\label{fig:mobilevit_fixed_profile_confusions}
\vspace{-.5em}
\end{figure*}

\subsubsection{Adaptive Profile Selection}
All controller settings are fixed using validation data before the test results reported here are evaluated. Building on the fixed-profile baselines above, Table~\ref{tab:fixed_cpi_main_comparison} separates the effects of adaptive profile selection and semantic early exit. Adaptive profiles without early exit observe the complete sequence, with $\overline{R}_{\rm obs}=100\%$, but reduce the average profile cost and normalized cumulative sensing cost to 0.852. In comparison with full-sequence High, this setting still achieves 86.76\% Macro-F1 at 14.81\% lower cost. With $\overline{C}=0.851$ and a 99.87\% observation ratio, the high-F1 Agentic setting achieves 86.75\% Macro-F1, and most of its resource reduction comes from profile adaptation.

\subsubsection{Semantic Early Exit}
With the sensing profile fixed at High, semantic early exit at $\tau=-0.05$ retains 90.16\% Macro-F1, only 0.54 percentage points below full-sequence High, while reducing the observation ratio and normalized cumulative cost from 100\% and 1.000 to 88.51\% and 0.885. Since $\overline{P}=1$, the savings arise entirely from shorter observation duration. This controlled comparison confirms that semantic early exit alone can reduce sensing use with a small recognition loss. The Agentic SemS reference further combines profile adaptation with early exit at $\tau=0$. Under the same $\lambda_{\rm cost}=0.005$ checkpoint, early exit reduces the observation ratio from 100\% to 86.74\% and cost from 0.852 to 0.747, a further 12.35\% relative reduction, while Macro-F1 decreases by only 0.97 percentage points from 86.76\% to 85.79\%. In comparison with full-sequence High, the complete controller reduces cost by 25.33\% while achieving 85.79\% Macro-F1. Thus, adaptation lowers acquisition cost, whereas early exit removes low-benefit observations.

\subsubsection{Per-Gesture Results}
\label{subsubsec:per_gesture_resource_allocation}
Table~\ref{tab:per_gesture_resource_allocation} examines the recognition--cost tradeoff of the resource-efficient Agentic SemS reference setting across gesture classes. In comparison with full-sequence High, cumulative sensing cost is reduced for every class, with reductions ranging from 12.36\% for Clap to 33.72\% for Sweep. The corresponding per-class F1 gaps range from 2.46 to 7.76 percentage points. These differences reflect the joint effects of adaptive profile selection and semantic early exit on the observation trajectories of individual gestures. Ground-truth labels are used only to compute the reported per-class statistics and remain unavailable to the online controller.

\begin{table}[t]
\centering
\caption{Per-class F1 and normalized cumulative sensing cost for full-sequence High and the Agentic SemS reference setting.}
\label{tab:per_gesture_resource_allocation}
\fontsize{8}{10}\selectfont
\renewcommand{\arraystretch}{1.10}
\setlength{\tabcolsep}{1.5pt}
\begin{tabular}{c|c|c|c|c|c}
\Xhline{1pt}
\multirow{2}{*}{Gesture}
& Full-seq.
& Agentic
& \multirow{2}{*}{$\Delta$F1 (pp)}
& \multirow{2}{*}{$\overline{C}$}
& Cost reduction \\
& High F1 (\%)
& reference (\%)
&
&
& vs. High (\%) \\
\Xhline{1pt}
Clap & 92.00 & 89.54 & $-2.46$ & 0.876 & 12.36 \\
\hline
Draw-O & 90.48 & 85.23 & $-5.25$ & 0.775 & 22.53 \\
\hline
Draw-Zigzag & 94.95 & 90.74 & $-4.21$ & 0.674 & 32.57 \\
\hline
Push \& Pull & 90.15 & 85.35 & $-4.80$ & 0.783 & 21.68 \\
\hline
Slide & 87.01 & 82.04 & $-4.97$ & 0.734 & 26.60 \\
\hline
Sweep & 89.59 & 81.83 & $-7.76$ & 0.663 & 33.72 \\
\hline
Overall Macro
& 90.70
& 85.79
& $-4.91$
& 0.747
& 25.33 \\
\Xhline{1pt}
\end{tabular}
\vspace{-1.5em}
\end{table}

\subsection{Belief Network and Motivation for Adaptive Control}
\label{subsec:belief_network_adaptive_control}

\subsubsection{Belief-Network Comparison}
Under the High profile, we compare independently trained MobileViT-XXS~\cite{mehta2022mobilevit}, gated recurrent unit (GRU)~\cite{cho2014learning}, and KV-cached causal Transformer models to determine a suitable architecture for the belief network. For this comparison, all models use the same 49,718-record Widar3.0 subset with a 70/30 train/test split. Fig.~\ref{fig:belief_network_comparison} reports Macro-F1 values of 91.93\%, 77.40\%, and 91.29\%, respectively. MobileViT-XXS treats each pre-segmented record as a single noncausal input, whereas the GRU and causal Transformer process it as an ordered sequence of CPI observations. The causal Transformer approaches the performance of MobileViT-XXS while supporting causal prefix processing and incremental KV-cached inference required for closed-loop sensing control. Agentic SemS therefore employs the causal Transformer as its belief network.

\begin{figure*}[t]
\centering
\begin{minipage}[t]{0.38\textwidth}
\centering
\includegraphics[width=\linewidth]{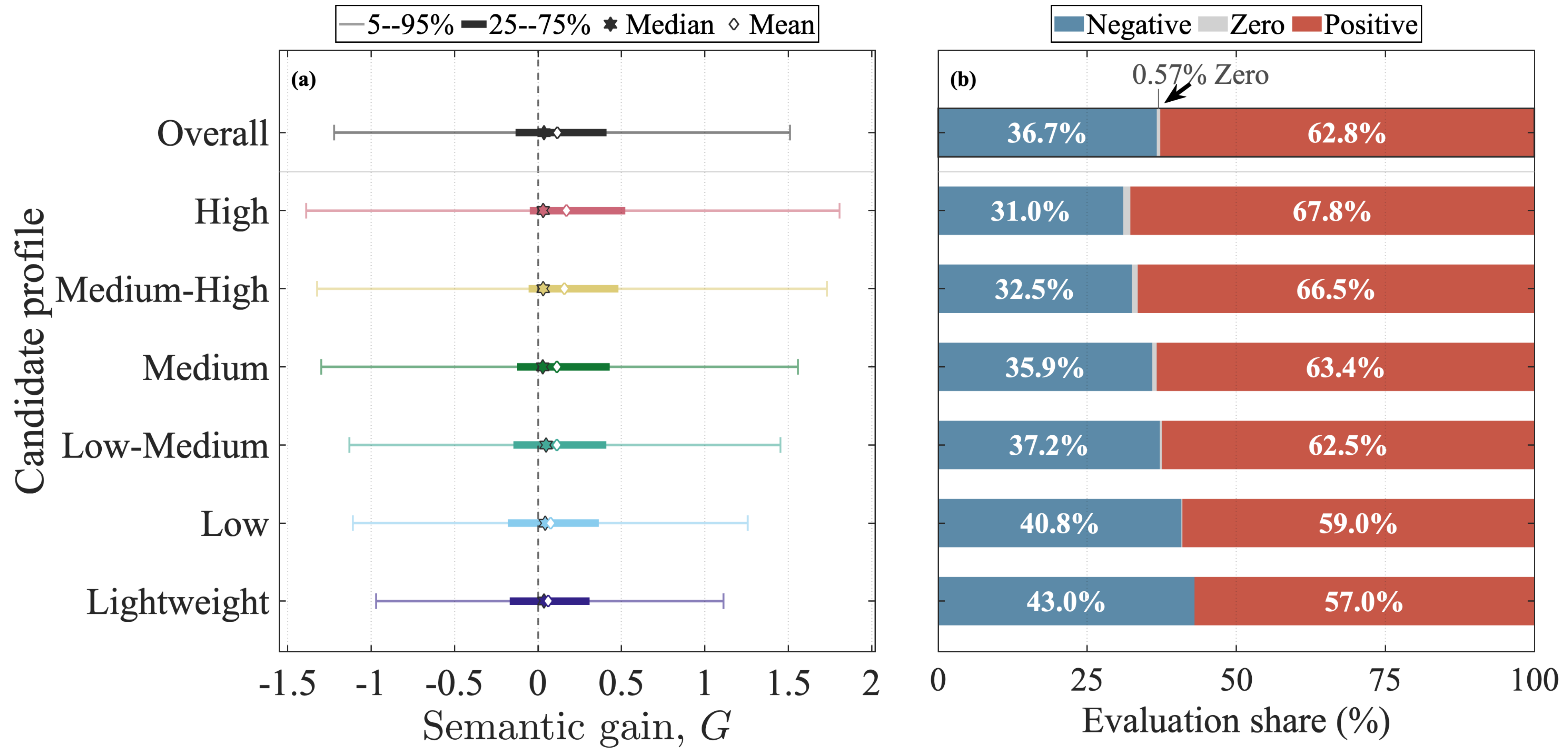}
\caption{Semantic-gain distributions across candidate profiles.}
\label{fig:semantic_gain_distribution}
\end{minipage}
\begin{minipage}[t]{0.3025\textwidth}
\centering
\includegraphics[width=\linewidth]{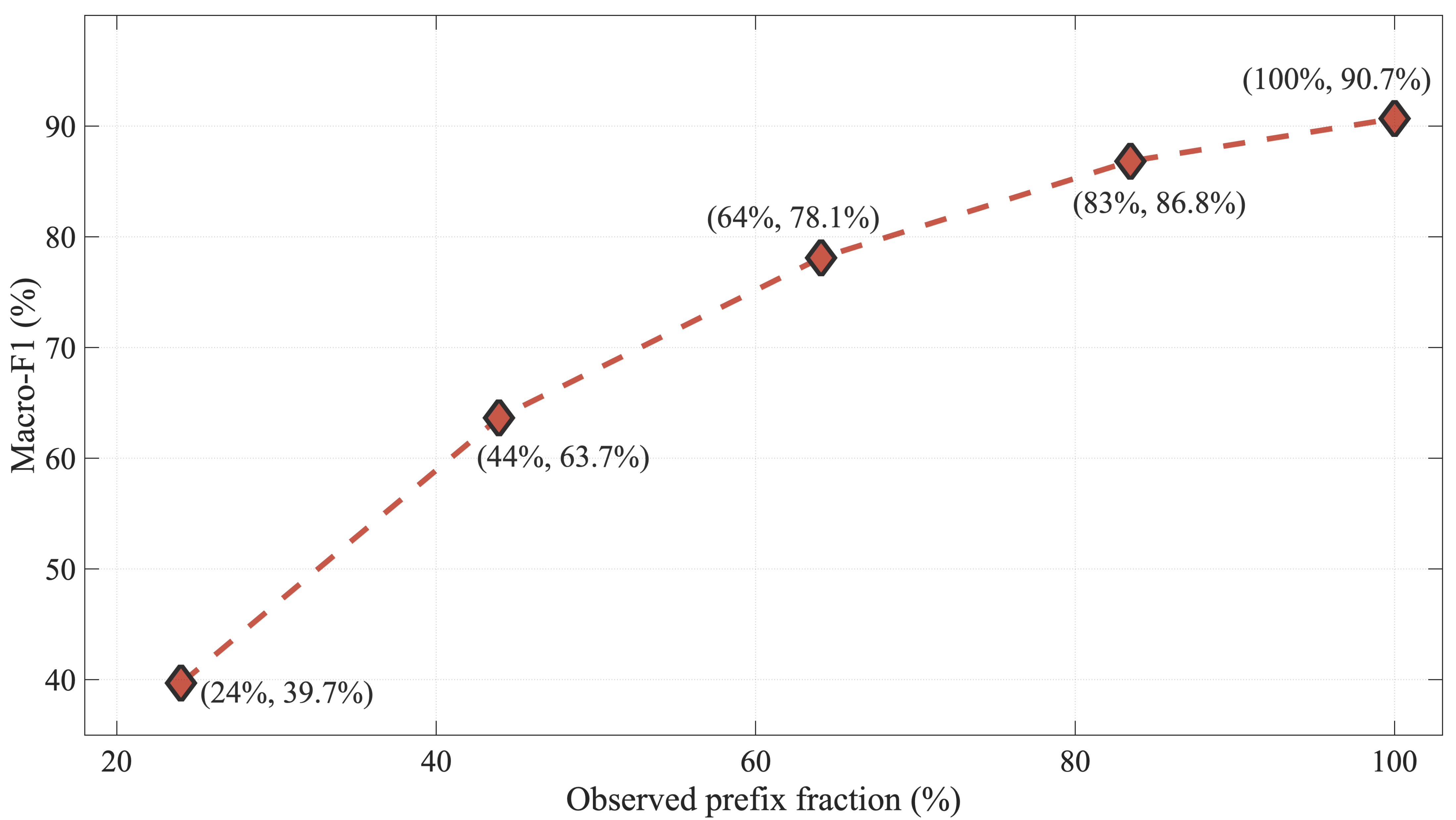}
\caption{Macro-F1 versus observation fraction.}
\label{fig:prefix_recognition}
\end{minipage}
\begin{minipage}[t]{0.3025\textwidth}
\centering
\includegraphics[width=\linewidth]{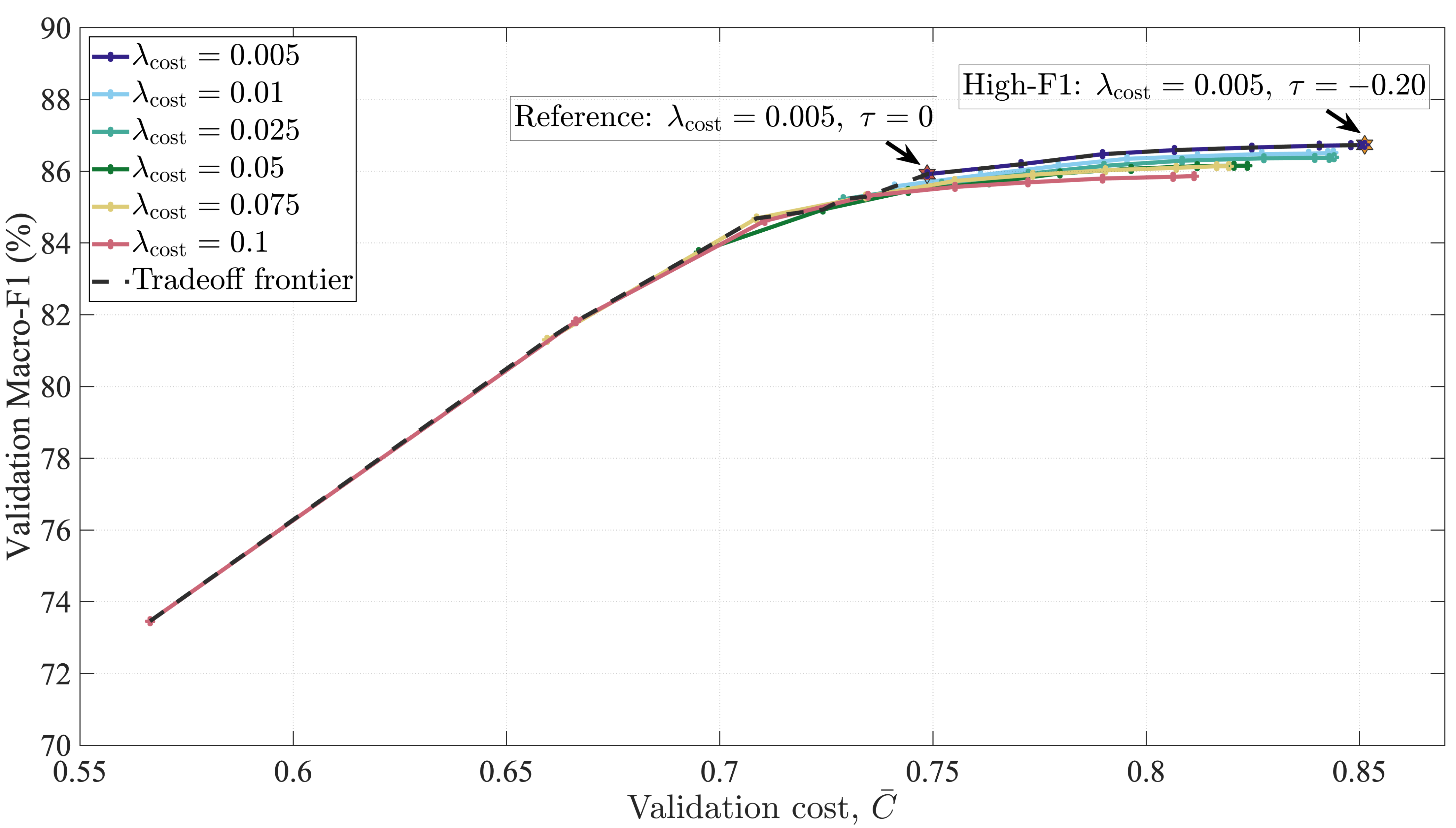}
\caption{Validation tradeoff across $\lambda_{\rm cost}$ and $\tau$.}
\label{fig:controller_validation_frontier}
\end{minipage}
\vspace{-1.5em}
\end{figure*}

\subsubsection{Profile-Dependent Recognition}
Fig.~\ref{fig:mobilevit_fixed_profile_confusions} presents the complete-sequence MobileViT-XXS confusion matrices across the six sensing profiles, illustrating how class separability changes with the available sensing information. Recognition generally improves from Lightweight to High, although some gesture classes retain strong recognition under lower-resource profiles. The highest-resource profile is therefore not always necessary, motivating profile adaptation according to the evolving semantic evidence. The online Agentic SemS controller uses the causal Transformer selected above.

\subsubsection{Semantic-Gain Analysis}
Fig.~\ref{fig:semantic_gain_distribution} examines the semantic gains used to construct the utility targets. The left plot shows the gain distributions for the six profiles and their aggregate, while the right plot reports the proportions of negative, numerically zero, and positive gains. Across 29,941 candidate evaluations from 1,000 deterministic class-stratified validation records, 62.77\% of the realized gains are positive, 36.66\% are negative, and 0.57\% are numerically zero. The overall mean and median are 0.115 and 0.036, respectively. A positive gain indicates that the next observation block increases the probability assigned to the correct class, whereas a negative gain indicates that the updated belief assigns it a lower probability. These gains characterize individual continuations rather than the intrinsic quality of a sensing profile. The same profile can therefore produce different gains depending on the current semantic state and the acquired observation. Negative gains for individual observations do not contradict the nonnegative expected VoI in Section~\ref{sec:voi_interpretation}, since the latter is defined as an expectation over possible future observations. The observed variation motivates candidate-conditioned utility prediction rather than fixed profile-upgrade decisions.

\begin{table}[t]
\centering
\caption{Validation sensitivity to the minimum observation budget for fixed $\lambda_{\rm cost}=0.005$ and $\tau=0$.}
\label{tab:nmin_sensitivity}
\fontsize{8}{10}\selectfont
\renewcommand{\arraystretch}{1.12}
\setlength{\tabcolsep}{3pt}
\resizebox{\columnwidth}{!}{
\begin{tabular}{c|c|c|c|c|c}
\Xhline{1pt}
$N_{\min}$ (CPIs) & Macro-F1 (\%) & $\overline{R}_{\rm obs}$ (\%) & $\overline{C}$ & Exit rate (\%) & Average exit CPI \\
\Xhline{1pt}
4  & 22.22 & 12.03 & 0.0899 & 97.50 & 4.60 \\
\hline
8  & 85.92 & 87.03 & 0.7487 & 37.93 & 35.87 \\
\hline
12 & 85.92 & 87.03 & 0.7487 & 37.93 & 35.87 \\
\hline
16 & 85.94 & 87.11 & 0.7495 & 37.88 & 35.96 \\
\Xhline{1pt}
\end{tabular}
}
\vspace{-1.5em}
\end{table}

\subsubsection{Recognition versus Observed Prefix}
Fig.~\ref{fig:prefix_recognition} shows how test-set recognition evolves as more of a High-profile sequence is observed. The 20\%, 40\%, 60\%, 80\%, and 100\% prefix settings correspond to average realized observation fractions of 24.03\%, 43.94\%, 64.09\%, 83.46\%, and 100\%, respectively, yielding Macro-F1 values of 39.68\%, 63.66\%, 78.12\%, 86.85\%, and 90.70\%. Improvements are largest early and diminish near the complete sequence, showing that additional sensing becomes less valuable as evidence accumulates and motivating continuation decisions from the current belief state.

\begin{table}[t]
\centering
\caption{Validation effect of $\tau$ for the fixed $\lambda_{\rm cost}=0.005$ utility checkpoint with $N_{\min}=8$ CPIs.}
\label{tab:resource_control_decomposition}
\fontsize{8}{10}\selectfont
\renewcommand{\arraystretch}{1.12}
\setlength{\tabcolsep}{3pt}
\begin{tabular}{c|c|c|c|c|c}
\Xhline{1pt}
Exit setting & Macro-F1 (\%) & $\overline{P}$ & $\overline{R}_{\rm obs}$ (\%) & $\overline{C}$ & Exit rate (\%) \\
\Xhline{1pt}
$\tau=-0.20$ & 86.74 & 0.852 & 99.89 & 0.851 & 0.41 \\
\hline
$\tau=-0.15$ & 86.72 & 0.854 & 99.35 & 0.848 & 2.27 \\
\hline
$\tau=-0.10$ & 86.66 & 0.856 & 96.62 & 0.825 & 12.39 \\
\hline
$\tau=-0.05$ & 86.48 & 0.860 & 92.31 & 0.790 & 25.16 \\
\hline
$\tau=-0.025$ & 86.19 & 0.862 & 89.88 & 0.771 & 31.31 \\
\hline
$\tau=0$ & 85.92 & 0.865 & 87.03 & 0.749 & 37.93 \\
\Xhline{1pt}
\end{tabular}
\vspace{-1.5em}
\end{table}

\subsection{Parameter Selection and Early Exit}
\label{subsec:controller_sensitivity}
\subsubsection{Role of the Minimum Observation Budget}
The minimum observation budget $N_{\min}$ and the early-exit threshold $\tau$ govern complementary parts of stopping. The former determines the earliest decision epoch at which semantic early exit is permitted, whereas the latter determines whether the predicted continuation utility is sufficiently low to stop thereafter. Table~\ref{tab:nmin_sensitivity} evaluates $N_{\min}$ on all 16,572 validation records with $\lambda_{\rm cost}=0.005$ and $\tau=0$. Allowing semantic early exit at the first utility evaluation with $N_{\min}=4$ causes 97.50\% of the records to exit early and reduces Macro-F1 to 22.22\%. This shows that four CPIs often provide insufficient semantic evidence for reliable stopping, causing premature exits before the belief becomes informative. Increasing $N_{\min}$ to 8 CPIs raises Macro-F1 to 85.92\% with a normalized cumulative cost of 0.7487. Results at 8 and 12 CPIs are identical, while increasing the budget to 16 CPIs changes Macro-F1 by only 0.02 percentage points and cumulative cost by less than 0.001. We therefore select $N_{\min}=8$ as the smallest tested budget that avoids premature stopping at the first decision epoch while retaining later early-exit opportunities. This budget covers two four-CPI observation blocks, allowing the controller to adapt the sensing profile after CPI 4 and evaluate semantic early exit from CPI 8 onward. Since $N_{\min}$ is applied only during inference, changing it does not require retraining either network. All exit-enabled comparisons therefore use $N_{\min}=8$.

\begin{table}[t]
\centering
\caption{Early-exit settings under fixed profiles on the test set.}
\label{tab:fixed_profile_early_exit}
\fontsize{8}{10}\selectfont
\renewcommand{\arraystretch}{1.10}
\setlength{\tabcolsep}{3pt}
\begin{tabular}{c|c|c|c|c|c}
\Xhline{1pt}
Profile & Exit setting & $\tau$ & Macro-F1 (\%) & $\overline{R}_{\rm obs}$ (\%) & $\overline{C}$ \\
\Xhline{1pt}
High & Disabled & -- & 90.70 & 100.00 & 1.000 \\
\hline
High & Higher-F1 & $-0.15$ & 90.60 & 98.45 & 0.985 \\
\hline
High & Lower-cost & $-0.05$ & 90.16 & 88.51 & 0.885 \\
\hline
Medium-High & Disabled & -- & 87.62 & 100.00 & 0.936 \\
\hline
Medium-High & Higher-F1 & $-0.125$ & 87.48 & 97.45 & 0.912 \\
\hline
Medium-High & Lower-cost & $-0.025$ & 86.80 & 86.90 & 0.813 \\
\Xhline{1pt}
\end{tabular}
\vspace{-1.5em}
\end{table}

\subsubsection{Validation-Set Search over $\lambda_{\rm cost}$ and $\tau$}
The validation search evaluates six separately trained utility checkpoints using $\lambda_{\rm cost}=0.005$, $0.01$, and $0.025$--$0.10$ in increments of $0.025$, with nine $\tau$ values from $-0.20$ to $0$ in increments of $0.025$. Because $\lambda_{\rm cost}$ changes the supervised utility target, each value requires a separate checkpoint, whereas $\tau$ varies during inference without retraining. Fig.~\ref{fig:controller_validation_frontier} summarizes all 54 combinations: each curve represents one checkpoint, its markers denote different $\tau$ values, and the dashed line connects the best observed recognition--cost tradeoffs. Increasing $\tau$ generally increases early stopping and lowers cost and Macro-F1. The highest validation Macro-F1, 86.74\%, occurs at $\lambda_{\rm cost}=0.005$ and $\tau=-0.20$, defining the high-F1 setting. At this checkpoint, $\tau=0$ defines the resource-efficient reference by lowering normalized cumulative cost from 0.851 to 0.749.

\subsubsection{Effect of the Early-Exit Threshold}
Table~\ref{tab:resource_control_decomposition} fixes $\lambda_{\rm cost}=0.005$ and $N_{\min}=8$ while varying $\tau$ to isolate semantic early exit. As $\tau$ increases from $-0.20$ to $0$, the exit rate rises from 0.41\% to 37.93\%, the observation ratio falls from 99.89\% to 87.03\%, and normalized cumulative cost falls from 0.851 to 0.749. Macro-F1 decreases from 86.74\% to 85.92\%. Thus, $\tau$ yields a 12.0\% relative validation-cost reduction for a 0.82-percentage-point Macro-F1 decrease. The smaller change in average profile cost indicates that $\tau$ primarily controls sensing duration rather than profile selection.

\begin{table}[t]
\centering
\caption{Sensitivity to the initial sensing profile on the test set.}
\label{tab:initial_profile_sensitivity}
\fontsize{8}{10}\selectfont
\renewcommand{\arraystretch}{1.12}
\setlength{\tabcolsep}{5pt}
\begin{tabular}{c|c|c|c|c}
\Xhline{1pt}
Initial profile & Macro-F1 (\%) & $\overline{P}$ & $\overline{R}_{\rm obs}$ (\%) & $\overline{C}$ \\
\Xhline{1pt}
Lightweight & 80.09 & 0.719 & 91.65 & 0.661 \\
\hline
Random & 85.79 & 0.866 & 86.74 & 0.747 \\
\hline
High & 88.85 & 0.961 & 83.71 & 0.800 \\
\Xhline{1pt}
\end{tabular}
\vspace{-1.5em}
\end{table}

\subsection{Fixed-Profile Early-Exit Baselines}
Table~\ref{tab:fixed_profile_early_exit} isolates sensing-duration control by enabling semantic early exit while keeping the sensing profile fixed. All exit-enabled settings use $\lambda_{\rm cost}=0.005$ and $N_{\min}=8$. Under fixed High, the Higher-F1 setting achieves 90.60\% Macro-F1 with an observation ratio of 98.45\%, whereas the Lower-cost setting reduces the observation ratio to 88.51\% while retaining 90.16\% Macro-F1. A similar trend is observed under fixed Medium-High. Because the sensing profile remains fixed, the reduction in cumulative sensing cost is attributable entirely to shorter sensing duration. Under fixed High, the Lower-cost setting reduces cumulative cost by 11.49\% with a decrease of only 0.54 percentage points in Macro-F1. Under fixed Medium-High, the corresponding relative cost reduction is 13.14\% with a decrease of 0.82 percentage points. These controlled results show that semantic early exit reduces sensing-resource use independently of adaptive profile selection.

\begin{table}[t]
\centering
\caption{Next-profile probabilities on the test set, conditional on continuation, for the reference controller.}
\label{tab:profile_transition_probability}
\fontsize{7.5}{9}\selectfont
\renewcommand{\arraystretch}{1.20}
\setlength{\tabcolsep}{1.8pt}
\begin{tabular}{c|c|c|c|c|c|c}
\Xhline{1pt}
Current & Lightweight & Low & Low-Medium & Medium & Medium-High & High \\
\Xhline{1pt}
Lightweight & 39.5 & 60.5 & -- & -- & -- & -- \\
\hline
Low & 31.8 & 10.7 & 57.5 & -- & -- & -- \\
\hline
Low-Medium & -- & 20.9 & 2.8 & 76.4 & -- & -- \\
\hline
Medium & -- & -- & 19.9 & 9.0 & 71.1 & -- \\
\hline
Medium-High & -- & -- & -- & 24.0 & 5.3 & 70.7 \\
\hline
High & -- & -- & -- & -- & 9.9 & 90.1 \\
\Xhline{1pt}
\end{tabular}
\vspace{-1.5em}
\end{table}

\subsection{Controller Behavior}
\label{subsec:controller_behavior}
\subsubsection{Initial-Profile Sensitivity}
Table~\ref{tab:initial_profile_sensitivity} evaluates the sensitivity of the reference controller to its initial sensing profile. Starting from Lightweight produces the lowest cumulative sensing cost of 0.661, but also the lowest Macro-F1 of 80.09\%. Starting from High increases Macro-F1 to 88.85\% while increasing cumulative sensing cost to 0.800. The random initialization used in the main reference experiments lies between these cases, achieving 85.79\% Macro-F1 with $\overline{C}=0.747$. Because profile transitions are restricted to adjacent profiles and occur only at decision epochs, initialization affects early resource use and shifts the resulting recognition--cost operating point. Subsequent utility-based decisions can nevertheless revise the initial allocation throughout the observation sequence.

\subsubsection{Profile Transitions}
Table~\ref{tab:profile_transition_probability} reports the next-profile probabilities for continuation decisions of the reference controller, excluding stop decisions and record-end carryover. Across 304,371 continuation decisions, retaining the current profile, moving to a higher-resource profile, and moving to a lower-resource profile account for 54.45\%, 30.82\%, and 14.72\%, respectively. Although upward transitions are more frequent than downward transitions in aggregate, both directions occur throughout online sensing. At the interior profiles, the controller frequently transitions to an adjacent profile as the semantic state evolves, whereas High remains unchanged in 90.12\% of its continuation decisions. The presence of both upward and downward transitions shows that adaptation is not a one-way resource-reduction strategy. When the predicted continuation utility favors additional sensing resources, the controller can move to a higher-resource profile; at other decision epochs, it can move to a lower-resource profile to reduce sensing cost. Together with the stay decisions, these transitions demonstrate that sensing-resource allocation is repeatedly adjusted according to the evolving semantic state rather than following a predetermined profile schedule.

\section{Conclusion}
Agentic SemS is a closed-loop framework that adapts sensing profiles and observation duration to evolving semantic evidence. A profile-conditioned causal Transformer maintains semantic belief, while a candidate-conditioned utility network evaluates profiles. Adaptive profile selection controls acquisition cost, whereas semantic early exit controls sensing duration. Widar3.0 experiments demonstrate their complementary roles. In comparison with full-sequence High, the resource-efficient Agentic setting achieves 85.79\% Macro-F1 while reducing normalized cumulative sensing cost by 25.33\%. Adaptive profile selection alone reduces cumulative sensing cost by 14.81\%, and semantic early exit at the same utility checkpoint provides a further 12.35\% relative reduction over adaptive sensing without early exit, with Macro-F1 decreasing by only 0.97 percentage points. Semantic feedback thus jointly controls sensing configuration and duration. Future work will dynamically update the communication-feasible profile set and evaluate end-to-end energy, latency, signaling overhead, and control overhead on physical AI-RAN platforms.

\ifCLASSOPTIONcaptionsoff
  \newpage
\fi

\bibliographystyle{IEEEtran}
\bibliography{main}

\end{document}